%% file: main.tex
\documentclass[numbers]{article}
\usepackage[preprint]{neurips_2026}

\usepackage{hyperref}
\usepackage{url}
\usepackage{amsmath,amssymb}
\usepackage{tikz-cd}
\usepackage{multicol}
\usepackage{algorithm}
\usepackage{algpseudocode}
\usepackage{float}
\usepackage{graphicx}
\usepackage{array}
\usepackage{xspace}
\usepackage{colortbl}
\usepackage{capt-of}
\usepackage{booktabs}
\usepackage{multirow}
\newcolumntype{C}[1]{>{\centering\arraybackslash}m{#1}}

\usepackage{natbib}

\title{Towards General Preference Alignment: Diffusion Models at Nash Equilibrium}

\author{%
Jiaming Hu\thanks{Co-first author.} \\
Boston University \\
\texttt{jh7453@bu.edu}
\And
Jiamu Bai\footnotemark[1] \\
Penn State University \\
\texttt{jvb6867@psu.edu}
\And
Haoyu Wang \\
University at Albany \\
\texttt{hwang28@albany.edu}
\And
Debarghya Mukherjee \\
Boston University \\
\texttt{mdeb@bu.edu}
\And
Ioannis Ch. Paschalidis \\
Boston University \\
\texttt{yannisp@bu.edu}
}

\input{notations}

\begin{document}

\maketitle

\begin{abstract}
Reinforcement learning from human feedback (RLHF) has been popular for aligning text-to-image (T2I) diffusion models with human preferences. 
 As a mainstream branch of RLHF, Direct Preference Optimization (DPO) offers a computationally efficient alternative that avoids explicit reward modeling and has been widely adopted in diffusion alignment. However, existing preference-based methods for diffusion alignment still rely on reward-induced preference signals and typically assume that human preferences can be adequately modeled by the Bradley--Terry (BT) model, which may fail to capture the full complexity of human preferences.
 In this paper, we formulate diffusion alignment from a game-theoretic perspective. We propose Diffusion Nash Preference Optimization (\ourmech), an intuitive general preference framework for diffusion alignment. 
\ourmech encourages the current policy to play against itself to achieve self-improvement and lead to a better alignment. 
Empirically, we demonstrate the effectiveness of \ourmech on the text-to-image generation task via various metrics. \ourmech consistently outperforms existing preference-based diffusion alignment methods.
\end{abstract}

\section{Introduction}
\input{1_intro.tex}
\section{Preliminaries}
\input{3_preliminary.tex}

\section{NLHF for Diffusion Models}
\input{4_algo.tex}

\section{Experiments}
\input{5_experiment.tex}

\section{Conclusion}
In this paper, we proposed Diffusion Nash Preference Optimization (\ourmech), a general preference optimization framework for diffusion models derived from a nash learning perspective. By directly optimizing relative preference probabilities through a self-play-style objective with reference-policy regularization, \ourmech provides a simple and stable approach to online diffusion alignment. Experiments on Stable Diffusion 1.5 and SDXL across Pick-a-Pic, Parti-Prompts, and HPSV2 show that \ourmech consistently improves alignment over strong preference-optimization baselines. These results suggest that Nash-style online learning is a promising alternative to reward-centric formulations for diffusion-model alignment. One limitation of the current study is that our average-rank preference construction relies on five scoring models, and incorporating a broader set of scoring models may provide a more comprehensive preference signal. In future work, we plan to extend this framework to stronger online data-collection strategies, and broader diffusion generation and editing tasks.
\bibliographystyle{plainnat}
\bibliography{ref}
\clearpage
\appendix
\input{appendix.tex}

\end{document}

%% file: notations.tex
\newcommand{\ourmech}{\ifmmode\mathrm{Diff.\mbox{-}NPO}\else Diff.-NPO\fi\xspace}

\newcommand{\rvx}{\mathbf{x}}
\newcommand{\rvy}{\mathbf{y}}
\newcommand{\rvc}{\mathbf{c}}

%% file: 1_intro.tex
Diffusion models have become a dominant framework for text-to-image generation due to their ability to produce high-fidelity images that are well aligned with textual descriptions. Representative systems such as Stable Diffusion~\citep{SDXL, SD15}, GPT Image 2~\citep{openai2026gptimage2}, and DALL-E 2~\citep{ramesh2022hierarchical} exemplify the strong generative capability of this class of models. Although large-scale pretraining on web data provides a powerful foundation, it does not by itself guarantee outputs that adequately reflect human preference. Drawing inspiration from alignment methods developed for large language models, recent work has adapted preference-based objectives such as DPO to diffusion models~\citep{rafailov2024directpreferenceoptimizationlanguage,li2024aligning,yang2024using} , enabling learning from preference data without explicit reward modeling. By incorporating preference signals directly into the optimization process, these methods have demonstrated clear advantages over standard supervised fine-tuning~\citep{diffusiondpo, D3PO}.

Despite these advantages, most existing DPO-style diffusion alignment methods are inherited from the RLHF paradigm, where preference learning is often interpreted through an implicit reward induced by Bradley--Terry \citep{bradley1952rank} or Plackett--Luce (PL)\citep{plackett1975analysis} models. 
We argue that the BT-based preference modeling fails to characterize the complexity of human preferences. Human preferences can be intransitive as in rock--paper--scissors: rock beats scissors, scissors beats paper, and paper beats rock. Such cyclic preferences cannot be represented by a single transitive ranking. In contrast, the BT model assumes transitivity, implying that if A is preferred to B and B is preferred to C, then A must be preferred to C. Therefore, this assumption may not hold across diverse human groups and can contradict observed human decision-making \citep{may1954intransitivity,tversky1969intransitivity}.
These limitations suggest that, rather than first converting human preference into a reward-induced preference function by BT model, it may be preferable to directly model and optimize the general preference relation itself.


Motivated by this observation, we propose \ourmech, a diffusion alignment method from the perspective of Nash Learning from Human Feedback (NLHF)\citep{munos2024nash}. Instead of viewing alignment as the maximization of a BT-induced likelihood function\citep{diffusiondpo}, \ourmech is derived from a minimax-style game in which one policy attempts to generate samples preferred over those of another policy, leading to the direct optimization of relative preference probabilities. The quality of a response is characterized by its expected preference win rate against samples drawn from the policy itself, as judged by the preference function. Thus, responses are favored when they are more likely to be preferred over the policy's own typical outputs. Concretely, \ourmech can be interpreted as a game between the current policy and the previous policy. The current policy is optimized to increase the preference probability assigned to winning samples relative to losing samples, while the previous policy serves as an evolving opponent and the reference policy regularizes deviations of the current policy from the pretrained distribution.  A Nash equilibrium is reached when both players adopt a policy $\pi^{*}$
 such that no competing policy can achieve a higher win rate over it.

Our objective brings three advantages. First, \ourmech provides a \textit{general preference perspective} on preference alignment. Instead of interpreting pairwise preferences through the combination of implicit rewards and BT model, as in standard DPO-style derivations, \ourmech directly optimizes preference relations. Second, \ourmech yields a \textit{self-play alignment objective} that connects DPO-style alignment and self-play within a unified formulation through the regularization term. In particular, different values of the regularization coefficient interpolate between these two regimes. Third, \ourmech makes the modeling assumptions behind preference optimization explicit. Under our formulation, the objective can be interpreted as fitting a preference-induced target distribution constructed from three components: the reference policy, the previous policy, and the preference signal. The regularization term determines how much this target distribution relies on the reference policy versus the previous policy. 

To evaluate the effectiveness of our method, we train Stable Diffusion 1.5 and SDXL on Pick-a-Pic \citep{kirstain2023pickapicopendatasetuser}, a large-scale dataset of real-world human preferences over image pairs. Empirical results show that \ourmech consistently improves alignment quality over existing preference-optimization baselines, demonstrating the effectiveness of on-policy Nash-style learning for diffusion model alignment. Our main contributions are as follows:

\begin{itemize}
    \item We bridge the gap between NLHF and diffusion model alignment by formulating diffusion alignment as a general preference game that avoids the Bradley--Terry assumption, thereby establishing a foundation for future research on Nash-style learning for diffusion models.
    \item We propose \ourmech, a novel on-policy general preference optimization algorithm for diffusion models. Our method unifies standard DPO and self-play paradigms, offering a general perspective for existing on-policy preference optimization methods.
    \item We empirically validate the effectiveness of \ourmech on text-to-image tasks. \ourmech achieves stronger alignment with human preferences than existing baselines.
\end{itemize}

%% file: 3_preliminary.tex
In this section, we introduce the basic background on diffusion models and Nash learning that will be used in our method, and defer a broader discussion of related work to Appendix~\ref{app:relatedwork}.

\subsection{Diffusion Models}
Diffusion models learn a data distribution by reversing a gradual noising process. 
The forward process starts from clean data $\rvx_0\sim p_{\mathrm{data}}$ and produces a sequence $\{\rvx_t\}_{t=1}^T$ by adding Gaussian noise with a schedule $\{\beta_t\}_{t=1}^T$ such that $0 \leq \beta_t \leq 1$.
Let $\alpha_t:=1-\beta_t$ and $\bar\alpha_t:=\prod_{s=1}^t \alpha_s$.
A common forward transition is
\begin{equation}
q(\rvx_t\mid \rvx_{t-1})=\mathcal{N}\bigl(\rvx_t; \sqrt{\alpha_t}\rvx_{t-1},(1-\alpha_t)I\bigr).
\end{equation}
The reverse (denoising) process is parameterized by a neural network $\epsilon_\theta$ that predicts the noise component given a noisy sample $\rvx_t$ at timestep $t$. Instead directly optimizing over $p_\theta(\rvx_{0})$, the diffusion model is trained to maximize Evidence Lower Bound (ELBO):
\begin{equation}
\label{eq:elbo}
\resizebox{0.93\textwidth}{!}{$
\begin{aligned}
  \log p_\theta(\rvx_{0}) \geq \mathbb{E}_{q(\rvx_{1:T}\mid \rvx_0)} \left[\log \frac{p_\theta(\rvx_{0:T})}{q(\rvx_{1:T}\mid \rvx_0)}\right] = \mathbb{E}_{q(\rvx_{1:T}\mid \rvx_0)} \left[\log \frac{p(\rvx_{T})\prod_{t=1}^T p_\theta(\rvx_{t-1} \mid \rvx_t)}{\prod_{t=1}^{T} q(\rvx_{t}\mid \rvx_{t-1})}\right],
\end{aligned}
$}
\end{equation}
which can be reduced to minimizing the denoising matching loss,
\begin{align*}
  \arg\min_{\theta} \mathrm{KL}\bigl(q(\rvx_{t-1}\mid \rvx_t, \rvx_0) \,\|\, p_\theta(\rvx_{t-1} \mid \rvx_t)\bigr)
  &= \arg\min_{\theta} \frac{(1-\alpha_t)^2}{2\sigma_t^2\alpha_t(1-\bar{\alpha}_t)}\,\mathbb{E}_{t, \rvx_0, \epsilon}\Big[\,\|\epsilon - \epsilon_\theta(\rvx_t, t)\|_2^2\Big].
\end{align*}
where $t\sim \mathcal{U}(0, T)$, and $\sigma_t^2=\frac{1-\bar{\alpha}_{t-1}}{1-\bar{\alpha}_t}\beta_t$, $\epsilon_\theta(\rvx_t, t)$ is the noise predicted by the diffusion model at timestep $t$, and $\epsilon\sim \mathcal{N}(0, I)$ is the Gaussian noise added to the image. 


\subsection{Diffusion Model Alignment with Human Feedback}

A common approach for aligning human preference for diffusion models is through DPO. Given a prompt \(\rvc \sim d_0\) and a clean image \(\rvx_0 \sim p_\theta(\cdot \mid \rvc)\) generated by the diffusion model, the standard RLHF objective can be written as
\begin{align}
  \min_{p_\theta} -\mathbb{E}_{p_\theta(\rvx_0\mid \rvc)} r(\rvc, \rvx_0) + \beta\mathrm{KL}(p_\theta(\rvx_0 \mid \rvc) ||p_\mathrm{ref}(\rvx_0 \mid \rvc)  ),
\end{align}
where $p_\theta$ and $p_{\mathrm{ref}}$ are probability distributions of the diffusion model and reference model, and $\beta$ controls the KL-regularization.
From ELBO trick in Equation~\ref{eq:elbo} and re-parameterization, the reward is calculated as,
\begin{align}
  r(\rvc, \rvx_0) = \beta \mathbb{E}_{p_\theta(\rvx_{1:T}\mid \rvx_0, \rvc)} \left[\log\frac{p_\theta(\rvx_{0:T}\mid \rvc)}{p_{\mathrm{ref}}(\rvx_{0:T}\mid \rvc)}\right] + \beta \log Z(\rvc),
\end{align}
where $Z(\rvc)$ is the partition function.
With BT model assumption and approximation of the reverse process, ~\citet{diffusiondpo} get rid of summing over probability from the whole chain and derive the objective below,
\begin{equation*}
    \mathcal{L}_{\text{Diff-DPO}}(\theta) 
    = - \mathbb{E}_{(\rvc, \rvx^+, \rvx^-)\sim \mathcal{D}, t} 
    \Big[ \log \sigma \big( \beta \, T \, \omega(\lambda_t) \, (\delta_\theta(\rvc,\rvx^+,t) - \delta_\theta(\rvc,\rvx^-,t)) \big) \Big],
\end{equation*}
where $\delta_\theta(\rvc, \rvx_t, t)
    := 
    -\Big(
    \|\epsilon - \epsilon_\theta(\rvx_t,\rvc,t)\|_2^2
    -
    \|\epsilon - \epsilon_{\mathrm{ref}}(\rvx_t,\rvc,t)\|_2^2
    \Big)$ 
can be seen as implicit reward, and $T$ is the number of diffusion steps and $\omega(\lambda_t)$ reweights timestep contributions. 

\subsection{Nash Learning with Human Feedback}
Many preference alignment methods rely on parametric preference models, such as Bradley--Terry, that represent pairwise preferences as differences in an underlying reward function\citep{bakker2022fine,jiang2025improving, wu2025rethinking}.
A key limitation is that such models impose transitivity and therefore cannot faithfully represent intransitive preferences that can arise in practice.
Instead, we directly work with the general preference probability $\mathbb{P}(\rvx_{0}\succ \rvx_{0}'\mid \rvc)$, 
and cast policy optimization as a two-player zero-sum game.

Given a prompt distribution $d_0$ and a reference policy $p_{\mathrm{ref}}$, for two policies $p_1$ (max-player) and $p_2$ (min-player) we define
\begin{equation}
\begin{aligned}
J(p_1,p_2)
&=\mathbb{E}_{\rvc\sim d_0}\Big[
  \mathbb{E}_{\rvx_{0}\sim p_1(\cdot\mid \rvc),\,\rvx_{0}'\sim p_2(\cdot\mid \rvc)}\big[\mathbb{P}(\rvx_{0}\succ \rvx_{0}'\mid \rvc)\big]\\
&\qquad\qquad
  -\tau\,\mathrm{KL}\big(p_1(\cdot\mid \rvc)\,\|\,p_{\mathrm{ref}}(\cdot\mid \rvc)\big)
  +\tau\,\mathrm{KL}\big(p_2(\cdot\mid \rvc)\,\|\,p_{\mathrm{ref}}(\cdot\mid \rvc)\big)
\Big].
\end{aligned}
\end{equation}
where $\tau>0$ controls how strongly both players are regularized toward $p_{\mathrm{ref}}$ via KL regularization.
Intuitively, $p_1$ tries to increase its win-rate against $p_2$ while staying close to the reference, and $p_2$ plays the adversary.

\subsubsection{Nash Policy and Duality Gap}
We restrict attention to a policy class $\mathcal{P}$ (e.g., policies with the same support as $p_{\mathrm{ref}}$).
A Nash equilibrium is defined by
\begin{equation}
p_1^*,p_2^*
\;:=\; \arg\max_{p_1\in\mathcal{P}}\;\arg\min_{p_2\in\mathcal{P}}\; J(p_1,p_2).
\end{equation}
In the symmetric setting, the two equilibrium policies coincide, and we denote the unique Nash policy by $p^*$.
For any candidate policy $p\in\mathcal{P}$, we measure how close it is to equilibrium via the duality gap
\begin{equation}
\mathrm{DualGap}(p)
\;:=\;
\max_{p_1\in\mathcal{P}} J(p_1,p)
\;-
\min_{p_2\in\mathcal{P}} J(p,p_2).
\end{equation}
By construction, $\mathrm{DualGap}(p)\ge 0$ and $\mathrm{DualGap}(p)=0$ if and only if $p$ is a Nash policy.
When $\mathrm{DualGap}(p)\le \varepsilon$, we call $p$ an $\varepsilon$-approximate Nash policy.

%% file: 4_algo.tex
\subsection{Overview}
Our method draws inspiration from Iterative Nash Policy Optimization (INPO)\citep{zhang2024iterative}, but replaces its fixed-margin regression loss with a pairwise logistic loss that encourages the model to assign a higher trajectory likelihood to the preferred image than to the rejected image. To make the Nash framework suitable for diffusion alignment, we follow the online mirror descent principle used in prior Nash-style alignment methods and derive a self-play objective tailored to diffusion models. We denote the policy after $s$ training steps by $p_s$.

\subsection{Online Mirror Descent for Nash Policy}

Given the preference probability $\mathbb{P}(\rvx_0\succ \rvx_0'\mid \rvc)$, define the loss at step $s$ as
\begin{align}
\ell_s(p)
&:=
-\mathbb{E}_{\rvc\sim d_0}\,\mathbb{E}_{\rvx_0\sim p(\rvx_0 \mid \rvc),\,\rvx_0'\sim p_s(\rvx_0'\mid \rvc)}\big[\mathbb{P}(\rvx_0\succ \rvx_0'\mid \rvc)\big]
+\tau\,\mathrm{KL}\big(p(\rvx_0\mid \rvc)\,\|\,p_{\mathrm{ref}}(\rvx_0\mid \rvc)\big) \notag \\
& \leq -\mathbb{E}_{\rvc\sim d_0}\,\mathbb{E}_{\rvx_{0:T}\sim p(\rvx_{0:T} \mid \rvc),\,\rvx_{0:T}'\sim p_s(\rvx_{0:T}\mid \rvc)}\big[\mathbb{P}(\rvx_{0:T} \succ \rvx_{0:T}'\mid \rvc)\big] \notag \\
& +\tau\,\mathrm{KL}\big(p(\rvx_{0:T}\mid \rvc)\,\|\,p_{\mathrm{ref}}(\rvx_{0:T}\mid \rvc)\big)
. 
\end{align}
Here we lift the image-level preference oracle to the reverse trajectory space by assuming that preferences depend only on the final samples.
The loss function corresponds to the game objective of the max-player with the min-player fixed to the current policy $p_s$. It consists of two parts: the negative win rate of $p$ against $p_s$ and a KL penalty term that keeps $p$ close to the reference policy $p_{\mathrm{ref}}$. A natural self-play strategy is to take the best response 
\begin{equation}
  p_{s+1}=\arg\min_{p\in\mathcal{P}}\,\ell_s(p),
\end{equation}
that is, the best response to $p_s$. However, this greedy update can be unstable: the new policy $p_{s+1}$ may deviate significantly from $p_s$, and such instability can lead to undesirable linear regret~\citep{lattimore2020bandit}. To obtain a more stable update, we instead adopt optimistic mirror descent with entropy regularization, also known as Hedge~\citep{freund1997decision}, and define $p_{s+1}$ as the minimizer of
\begin{equation}
\arg\min_{p\in\mathcal{P}}
\Big\langle \nabla \ell_s(p_s),\,p\Big\rangle
+\eta\,\mathrm{KL}(p\,\|\,p_s),
\end{equation}
where $\eta>0$ and
\begin{equation}
\nabla_{p_s(\rvx_{0:T})} \ell_s(p_s)
=
-\mathbb{E}_{\rvx_{0:T}'\sim p_s}\big[\mathbb{P}(\rvx_{0:T} \succ \rvx_{0:T}')\big]
+
\tau\left(
\log \frac{p_s(\rvx_{0:T})}{p_{\mathrm{ref}}(\rvx_{0:T})} + 1
\right).
\end{equation}
Compared with the greedy update, this objective introduces an additional KL penalty that keeps $p_{s+1}$ close to $p_s$, thereby encouraging a more stable learning dynamic.

\subsection{Nash Preference Optimizaton}
A closed-form solution of the update implied by the OMD objective can be written in exponential-weights form; the full derivation is provided in Appendix~\ref{app:math-analysis}.
\begin{equation}
p_{s+1}(\rvx_{0:T}\mid \rvc)
\propto
\exp\Big(\tfrac{1}{\eta}\,\mathbb{P}(\rvx_{0:T}\succ p_s\mid \rvc)\Big)
\,p_{\mathrm{ref}}(\rvx_{0:T}\mid \rvc)^{\tfrac{\tau}{\eta}}
\,p_s(\rvx_{0:T}\mid \rvc)^{1-\tfrac{\tau}{\eta}},
\end{equation}
where $\mathbb{P}(\rvx_{0:T}\succ p_s\mid \rvc)=\mathbb{E}_{\rvx_{0:T}'\sim p_s(\cdot\mid \rvc)}[\mathbb{P}(\rvx_{0:T}\succ \rvx_{0:T}'\mid \rvc)]$.
Directly computing the normalization is intractable, so we optimize pairwise log-ratios.
For a response pair $(\rvx_{0:T},\rvx_{0:T}')$, define $h_s(p; \rvc,\rvx_{0:T},\rvx_{0:T}')$ as
\begin{equation}\label{eq:nash-obj}
  \frac{\tau}{\eta}\left(
    \log\frac{p(\rvx_{0:T}\mid \rvc)}{p(\rvx_{0:T}'\mid \rvc)}
    -\log\frac{p_{\mathrm{ref}}(\rvx_{0:T}\mid \rvc)}{p_{\mathrm{ref}}(\rvx_{0:T}'\mid \rvc)}
  \right)
  +
  \frac{\eta-\tau}{\eta}\left(
    \log\frac{p(\rvx_{0:T}\mid \rvc)}{p(\rvx_{0:T}'\mid \rvc)}
    -\log\frac{p_s(\rvx_{0:T}\mid \rvc)}{p_s(\rvx_{0:T}'\mid \rvc)}
  \right).
\end{equation}
At optimum, $h_s(p_{s+1};\rvc,\rvx_{0:T},\rvx_{0:T}')$ should equal the preference-margin difference:
\begin{equation}\label{eq:pref-margin}
  h_s(p_{s+1}; \rvc,\rvx_{0:T},\rvx_{0:T}')
  =
  \frac{\mathbb{P}(\rvx_{0:T}\succ p_s\mid \rvc)-\mathbb{P}(\rvx_{0:T}'\succ p_s\mid \rvc)}{\eta}.
\end{equation}
Equation~\eqref{eq:pref-margin} shows that, 
at the OMD-induced optimum,
$h_s$ should match the relative preference advantage between two trajectories
against the same preference oracle. 
However, this target is difficult to compute exactly,
as it requires estimating the expected preference of each trajectory against
samples drawn from $p_s$. We therefore use observed preferred--rejected pairs
as supervision and optimize a pairwise logistic loss based on $h_s$. This loss
encourages $h_s(p;\rvc,\rvx_{0:T},\rvx_{0:T}')$ to have the correct preference
direction, serving as a tractable surrogate for matching the ideal
preference-advantage difference in Equation~(14).
Based on above observation, we define the loss function $L_s(p)$ as
\begin{equation}\label{eq:obj-chain}
\begin{aligned}
L_s(p)
:={}\;
\mathbb{E}_{\rvc\sim d_0}\,\mathbb{E}_{\rvx_{0:T},\rvx_{0:T}'\sim p_s(\cdot\mid \rvc)}\Big[
  - & \log \sigma\Big(
     \frac{\tau \beta}{\eta}\Big(
      \log\frac{p(\rvx_{0:T}\mid \rvc)}{p(\rvx_{0:T}'\mid \rvc)}
      -\log\frac{p_{\mathrm{ref}}(\rvx_{0:T}\mid \rvc)}{p_{\mathrm{ref}}(\rvx_{0:T}'\mid \rvc)}
    \Big) \\
    &
    + \frac{(\eta-\tau)\beta}{\eta}\Big(
      \log\frac{p(\rvx_{0:T}\mid \rvc)}{p(\rvx_{0:T}'\mid \rvc)}
      -\log\frac{p_s(\rvx_{0:T}\mid \rvc)}{p_s(\rvx_{0:T}'\mid \rvc)}
    \Big)
  \Big)
\Big],
\end{aligned}
\end{equation}
where $0\leq \tau / \eta \leq 1$. This objective function further clarifies how our method differs from Diffusion-DPO. Whereas Diffusion-DPO is designed for the offline setting and mainly contrasts the current policy with the reference policy $p_{\mathrm{ref}}$, our method is inherently online and follows a self-play scheme in which each update is evaluated against samples generated by the current policy itself. Under this view, the next policy $p_{s+1}$ should stay close to both $p_s$ and $p_{\mathrm{ref}}$, although these two regularization effects serve different roles: proximity to $p_s$ promotes stable online optimization, while proximity to $p_{\mathrm{ref}}$ acts as an anchor that mitigates reward hacking.

In the above derivation, $(\rvx_{0:T},\rvx'_{0:T})$ denotes a generic pair of
trajectories sampled from the previous policy. In practice, after sampling such
a pair, the preference oracle is used to determine which trajectory is preferred.
We therefore relabel the preferred trajectory as $\rvx^+_{0:T}$ and the rejected
trajectory as $\rvx^-_{0:T}$.
For diffusion models, we identify the generic policy $p$ with the trainable reverse process $p_\theta$, and write the previous policy from last step as $p_{\mathrm{prev}}$. Following the standard diffusion-DPO reduction from full-chain likelihoods to per-step reverse transitions, we obtain the final single-timestep objective, with details deferred to Appendix~\ref{app:per-step-obj}:
\begin{align}\label{eq:obj-final}
 L_{\ourmech}= &
\mathbb{E}_{\rvc, t,  \rvx_0^+,\rvx_0^-\sim p_{\mathrm{prev}}(\cdot\mid \rvc), 
\rvx_t^+,\rvx_{t-1}^+, 
\rvx_t^-, \rvx_{t-1}^-} \Big[ -\log \sigma \Big(
    \beta\log\frac{p_\theta(\rvx_{t-1}^+\mid \rvx_t^+, \rvc)}
    {p_\theta(\rvx^-_{t-1}\mid \rvx^-_t, \rvc)} \notag \\
    -& \frac{\tau \beta}{\eta} 
      \log\frac{p_{\mathrm{ref}}(\rvx_{t-1}^+\mid \rvx_t^+, \rvc)}
      {p_{\mathrm{ref}}(\rvx^-_{t-1}\mid \rvx^-_t, \rvc)}
     - \frac{(\eta-\tau)\beta}{\eta}
      \log\frac{p_{\mathrm{prev}}(\rvx_{t-1}^+\mid \rvx_t^+, \rvc)}
      {p_{\mathrm{prev}}(\rvx^-_{t-1}\mid \rvx^-_t, \rvc)}
  \Big)
\Big],  
\end{align}
where $t$ is the time step sampled from $\mathcal{U}(0,T)$. Using Gaussian reparameterization of the reverse process, the objective can be further written as
\begin{align*}
  L_{\ourmech}= & \mathbb{E}_{\rvc, t, 
    \rvx_0^+,\rvx_0^-\sim p_{\mathrm{prev}}(\cdot\mid \rvc), 
    \rvx_t^{\{+, -\}}\sim q(\rvx_t^{\{+, -\}}\mid \rvx_0^{\{+, -\}})} \Big[- \log \sigma \Big(
  \beta T \omega(\lambda_t) \big[
  \delta_\theta(\rvc, \rvx^+_t, \rvx^-_t) \notag \\
    &  
  - \tau/\eta \delta_{\mathrm{ref}}(\rvc, \rvx^+_t, \rvx^-_t) 
  - (\eta-\tau)/\eta \delta_{\mathrm{prev}}(\rvc, \rvx^+_t, \rvx^-_t)\big]
\Big)\Big], 
\end{align*}
where $\delta_\theta$ is defined as
\begin{align*}
  \delta_\theta(\rvc, \rvx^+_t, \rvx^-_t) = -\left[\|\epsilon^+ - \epsilon_{\theta}(\rvx_t^+, t, \rvc)\|_2^2
    - \|\epsilon^- - \epsilon_{\theta}(\rvx_t^-, t, \rvc)\|_2^2\right],
\end{align*}
and $\delta_\mathrm{prev}$ and $\delta_\mathrm{ref}$ are defined analogously by replacing $\epsilon_\theta$ with $\epsilon_{\mathrm{prev}}$ and $\epsilon_{\mathrm{ref}}$, respectively. The full algorithm for \ourmech can be found in Algorithm~\ref{algo:npo} in Appendix~\ref{app:algo}.

\subsection{Theoretical Insights}\label{sec:theoretical}

In this section, we provide an interpretation of our objective through a preference-induced conditional distribution. Inspired by the form of guidance-based methods, we consider the following target distribution:
\begin{align}\label{eq:assumption}
  p_{\theta^*}(\rvx_{0:T} \mid \rvc)
  \propto
  p_{\mathrm{ref}}(\rvx_{0:T} \mid \rvc)^{\gamma}
  p_{\mathrm{prev}}(\rvx_{0:T} \mid \rvc)^{1-\gamma}
  p(\rvy \mid \rvx_{0:T}, \rvc)^{1/\beta},
\end{align}
where $\theta^*$ denotes the ideal target policy induced by the conditional distribution $p(\rvy \mid \rvx_{0:T}, \rvc)$, and $\rvy$ is a latent variable representing the  preference outcome. For simplicity we set $\gamma=\tau/\eta$. The term $p(\rvy \mid \rvx_{0:T}, \rvc)$ captures how likely a sample is preferred under the oracle. Since directly accessing $p(\rvy \mid \rvx_{0:T}, \rvc)$ is intractable, we instead optimize relative preference likelihoods. Connecting this assumption with our training objective, we obtain:
\begin{align*}
  \mathcal{L}_{\ourmech}
  =& -
  \mathbb{E}_{(\rvc, \rvx_{0:T}^+, \rvx_{0:T}^-) \sim \mathcal{D}}
  \\
  & \Bigg[
  -\log \sigma \Bigg(
  \beta
  \log\frac{p_\theta(\rvx_{0:T}^+|\rvc)}{p_\theta(\rvx_{0:T}^-|\rvc)}-\beta \Bigg[
  \gamma\log\frac{p_{\mathrm{ref}}(\rvx_{0:T}^+|\rvc)}{p_{\mathrm{ref}}(\rvx_{0:T}^-|\rvc)}
  + (1-\gamma)\log\frac{p_{\mathrm{prev}}(\rvx_{0:T}^+|\rvc)}{p_{\mathrm{prev}}(\rvx_{0:T}^-|\rvc)}
  \Bigg]
  \Bigg)
  \Bigg]. 
\end{align*}


By absorbing normalization constants into the definition of $p(\rvy \mid \cdot)$, we can express the objective as:
\begin{align} \label{eq:preference-obj}
  \mathcal{L}_{\ourmech}
  =
  - \mathbb{E}
  \Big[
  \log \sigma\big(
  \log p(\rvy \mid \rvx_{0:T}^+, \rvc)
  -
  \log p(\rvy \mid \rvx_{0:T}^-, \rvc)
  \big)
  \Big].
\end{align}

\paragraph{A Unified View of Preference Optimization.}
Equation~\ref{eq:preference-obj} reveals that a wide class of preference optimization methods---including DPO, self-play methods, and our approach---can be viewed as optimizing the same fundamental pairwise logistic objective. The objective encourages the model to assign a larger preference-induced likelihood to the preferred sample than to the dispreferred sample. This view also provides a useful interpretation of $\beta$. In standard RLHF, $\beta$ is usually introduced as the coefficient of the KL regularization term. Under our formulation, $\beta$ can be interpreted as controlling the strength of the preference-induced conditional distribution in Equation~\ref{eq:assumption}. 

The key distinction among these methods therefore lies not only in the loss form, but in how the implicit target distribution is modeled. When $\gamma=1$, Equation~\ref{eq:assumption} reduces to
\[
p_{\theta^*}(\rvx_{0:T}\mid \rvc)
\propto
p_{\mathrm{ref}}(\rvx_{0:T}\mid \rvc)
p(\rvy\mid \rvx_{0:T},\rvc)^{1/\beta},
\]
which corresponds to the Diffusion-DPO-style assumption: the target distribution is obtained by preference-conditioning the reference policy. While this provides a stable anchor, it may limit learning because all improvement is measured relative to a fixed reference model. When $\gamma=0$, Equation~\ref{eq:assumption} instead becomes
\[
p_{\theta^*}(\rvx_{0:T}\mid \rvc)
\propto
p_{\mathrm{prev}}(\rvx_{0:T}\mid \rvc)
p(\rvy\mid \rvx_{0:T},\rvc)^{1/\beta},
\]
which corresponds to a pure self-play update. This forces each iteration to move beyond the previous policy and can provide a stronger online learning signal. However, without the reference policy as an anchor, the model may drift away from the pretrained distribution and become unstable. Our objective interpolates between these two extremes. The reference policy provides a stable distributional anchor, while the previous policy introduces an online self-improvement signal. As a result, the model is not restricted to merely improving over the fixed reference, nor is it forced into unconstrained self-play at every step. This interpolation mitigates the two-sided failure mode of pure DPO and pure self-play, yielding a more stable and expressive preference-induced target distribution.

\paragraph{Why Square Loss in INPO is Unstable.}

INPO-style objectives~\citep{zhang2024iterative} justify the squared loss under
a specific population assumption: after sampling a pair $\rvx,\rvx'\sim p_s$,
the ordered pair $(\rvx^+,\rvx^-)$ is drawn from the preference distribution
$\lambda_p(\rvx,\rvx')$, i.e., $(\rvx,\rvx')$ is selected with probability
$\mathbb{P}(\rvx\succ\rvx')$ and $(\rvx',\rvx)$ with probability
1-$\mathbb{P}(\rvx\succ\rvx')$. However, in most practical implementations,
ordered preference pairs are constructed by querying an oracle or a reward
model on the two samples and assigning the higher-scored sample as $\rvx^+$ and
the lower-scored sample as $\rvx^-$. Thus, the observed pair is deterministically
ordered by the oracle score, rather than sampled according to the probabilistic
preference distribution $\lambda_p$. In our online diffusion setting, this
discrepancy is more explicit: we generate multiple candidate images, rank them
using the preference oracle, and select the best-ranked image as $\rvx^+$ and
the worst-ranked image as $\rvx^-$. Since these ordered pairs do not follow the
preference distribution assumption,
directly applying the squared surrogate becomes less natural for our data
construction. We empirically validate this behavior in
Section~\ref{sec:ablation}.

%% file: 5_experiment.tex
\subsection{Experiment Setup}
\paragraph{Models, datasets, baselines, and evaluation metrics.}
We fine-tune two text-to-image diffusion backbones, Stable Diffusion 1.5 (SD1.5) and SDXL. The prompts of training data are drawn from the Pick-a-Pic v1 preference dataset~\citep{kirstain2023pickapicopendatasetuser}. We use prompts from the Pick-a-Pic test split~\citep{kirstain2023pickapicopendatasetuser}, Parti-Prompts~\citep{yu2022scaling}, and HPSV2~\citep{wu2023humanpreferencescorev2} for testing. 
We compare against several representative baselines, including online SFT trained on preferred images, online  Diffusion-DPO~\citep{diffusiondpo}, SPIN~\citep{SPIN} and SEPPO~\citep{SEPPO}, together with the original pretrained SD1.5 and SDXL models. We use the public available checkpoints for SPIN\footnote{https://huggingface.co/UCLA-AGI/SPIN-Diffusion-iter3} and SEPPO\footnote{https://huggingface.co/DwanZhang/SePPO} for evaluation.
For text-to-image alignment task, we adopt five widely used automatic metrics: PickScore~\citep{kirstain2023pickapicopendatasetuser}, HPSV2~\citep{wu2023humanpreferencescorev2}, CLIP~\citep{CLIP}, Aesthetic Score~\citep{AES}, and ImageReward~\citep{xu2023imagereward}, and compare the metric scores and winrates among different methods.
 More details are provided in Appendix~\ref{app:exp-details}.

\paragraph{Online Generation and Preference Oracle.} During online generation, we use 10 inference steps for efficiency, following prior work showing that a small number of denoising steps has little impact on optimization performance~\citep{diffusionNFT, flowgrpo}. 
For each prompt, we generate 8 candidate images and use automatic preference oracles to construct preference pairs. 
Specifically, we evaluate each candidate using five scoring models: PickScore~\citep{kirstain2023pickapicopendatasetuser}, HPSV2~\citep{wu2023humanpreferencescorev2}, CLIP~\citep{CLIP}, Aesthetic Score~\citep{AES}, and ImageReward~\citep{xu2023imagereward}. 
For each oracle, we rank the 8 candidates according to their scores, and then compute the average rank across the five oracles as the final preference score. 
The candidate with the best average rank is selected as the positive sample, while the candidate with the worst average rank is selected as the negative sample.


\input{table/sd15_result}

\subsection{General Evaluation: Text-to-Image Alignment}
We report text-to-image generation results on SD1.5 and SDXL in
Tables~\ref{tab:sd15-result} and~\ref{tab:sdxl-result}. \ourmech achieves the
highest average win rate on every evaluation dataset across both SD1.5 and
SDXL, showing consistent gains over the pairwise Online-Diff-DPO baseline.
The improvement is particularly strong on SDXL, where \ourmech improves the
average win rate by 7.0\% relatively on Parti-Prompts and remains the best
method on Pick-a-Pic and HPSV2. 

\input{table/sdxl_result}

\begin{figure}[h]
    \centering
    \includegraphics[width=\textwidth]{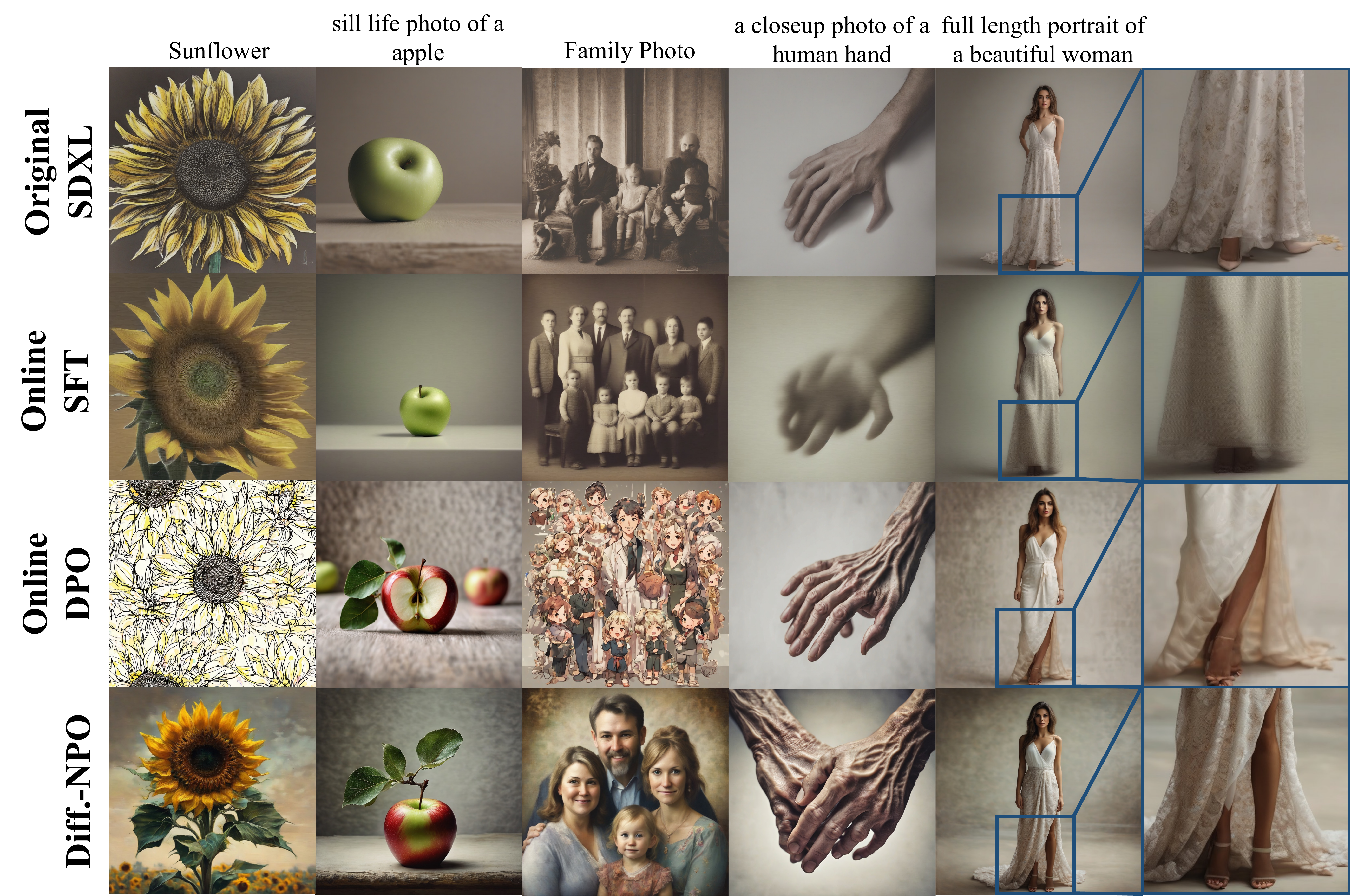}
    \vspace{-1.0em}
    \caption{Qualitative comparison on SDXL across six representative prompts. Rows correspond to the original SDXL model, SFT, Online DPO, and Diff.-NPO. Compared with the baselines, Diff.-NPO produces images that are generally more realistic, semantically faithful, and visually coherent.}
    \label{fig:sdxl-pics}
    
\end{figure}

The advantage of \ourmech is most evident on preference-oriented metrics.
On SDXL Parti-Prompts, \ourmech improves the ImageReward win rate from 79.88
to 86.73 and the AES win rate from 76.74 to 85.04, corresponding to relative
improvements of 8.6\% and 10.8\% over Online-Diff-DPO, respectively. 
In contrast, SFT is unstable and often performs
worse than the original SDXL model, with average win rates below 50\%. This is likely because SFT is trained directly on images generated with only 10 inference steps, which are often low quality, whereas RL-based methods such as \ourmech can leverage both positive and negative samples to improve even from poor initial generations.
Figure~\ref{fig:sdxl-pics} presents qualitative comparisons between \ourmech and the baselines after fine-tuning SDXL. Across diverse prompt types, including objects, human faces, multiple people, and full-body portraits, \ourmech consistently produces images with more coherent and  appealing visual quality. In particular, for the prompt ``full length portrait of a beautiful woman'', \ourmech better preserves anatomical details and generates clearer, more realistic full-body structures compared with the baselines. Qualitative results for SD1.5 and additional SDXL examples are provided in Figure~\ref{fig:sd15-pics} and Figure~\ref{fig:sdxl-pics-more}, respectively, in Appendix~\ref{app:more-pic-result}.

\begin{figure}[h!]
    \centering
    \begin{minipage}[t]{0.56\textwidth}
        \vspace{0pt}
        \centering
        \includegraphics[width=\textwidth]{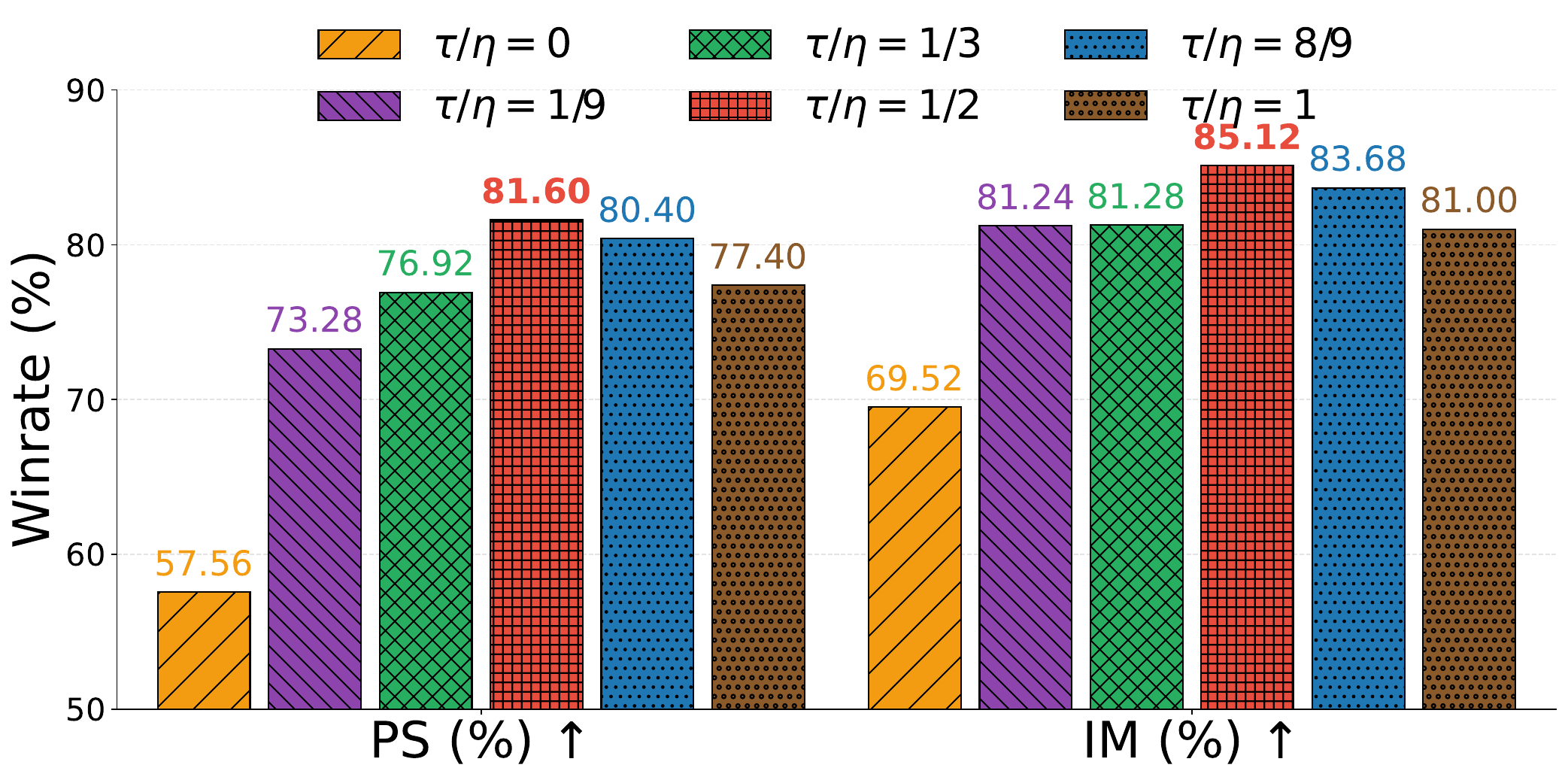}
        \vspace{-2.0em}
        \caption{Winrates of different $\tau/\eta$ values for PickScore and ImageReward. Full results in Figure~\ref{fig:ablation-all-5}, Appendix~\ref{app:reg-ablation}.}
        \label{fig:ablation-study}
    \end{minipage}\hfill
    \begin{minipage}[t]{0.40\textwidth}
        \vspace{0pt}
        \centering
        \captionof{table}{Ablation study of different $\tau/\eta$ settings on Pick-a-Pic for SD1.5, evaluated using automatic metrics. 
        }
        \label{tab:ablation-ratio}
        \vspace{0.3em}
        \input{table/ablation_ratio}
    \end{minipage}
\end{figure}

\subsection{Ablation Studies}\label{sec:ablation}

\paragraph{Impact of different $\tau/\eta$ regularization Coefficient.}
We study the effect of the regularization ratio $\tau/\eta$, which controls the relative strength of the reference-policy and previous-policy regularization terms. When $\tau/\eta=1$, the objective only regularizes toward the reference policy and reduces to the DPO-style setting; when $\tau/\eta=0$, the objective only regularizes toward the previous policy, corresponding to a pure self-play update. Table~\ref{tab:ablation-ratio} reports the automatic metric scores under different $\tau/\eta$ values, while Figure~\ref{fig:ablation-study} visualizes the win-rate comparison on PickScore and ImageReward. The full win-rate comparison across all five metrics is provided in Figure~\ref{fig:ablation-all-5} in Appendix~\ref{app:reg-ablation}. We observe a clear increase-then-decrease trend as $\tau/\eta$ varies from $0$ to $1$, with the best overall performance achieved at an intermediate value. This suggests that relying solely on either the reference policy or the previous policy is suboptimal, and highlights the importance of jointly using both policies to provide stable and effective regularization.

\begin{figure}[h]
    \centering
    \begin{minipage}[t]{0.57\textwidth}
        \vspace{6pt}
        \textbf{Impact of the objective design.} We further study the importance of the pairwise-logistic objective (PL), $\log \sigma(\cdot)$, by comparing it with the squared distance (SD) used in INPO-style objectives. As discussed in Section~\ref{sec:theoretical}, the squared distance objective relies on a preference distribution assumption, whereas practical implementations often construct pairs by deterministically ranking generated samples with an oracle. Table~\ref{tab:square-loss-ablation} shows that replacing the pairwise-logistic objective with squared distance leads to a substantial performance drop on win-rate evaluations, which supports our design of using the pairwise-logistic objective, which ensures stable optimization.
    \end{minipage}\hfill
    \begin{minipage}[t]{0.4\textwidth}
        \vspace{0pt}
        \centering
        \captionof{table}{Results on the objective design, comparing the squared distance objective (SD) with the pairwise-logistic objective (PL) used in \ourmech. }
        \label{tab:square-loss-ablation}
        \input{table/ablation_square_loss}
    \end{minipage}
\end{figure}
\vspace{-1.5em}

%% file: table/sd15_result.tex
\begin{table}[t]
\centering
\small
\setlength{\tabcolsep}{1.5pt}
\renewcommand{\arraystretch}{1.0}
\caption{SD1.5 win-rate and metric results for \ourmech compared with other baselines. Win rates are vs.~original SD1.5. The yellow-highlighted rows indicate our method. For each column, the best result is shown in \textbf{bold}, and the second-best result is \underline{underlined}.}
\label{tab:sd15-result}
\vspace{-0.75em}
\begin{tabular}{l l *{6}{C{2.75em}} *{3}{C{2.75em}} *{2}{C{3.45em}} }
\toprule 
\multirow{2}{*}{\textbf{Dataset}} & \multirow{2}{*}{\textbf{Method}} &
\multicolumn{6}{c}{\textbf{SD1.5 Winrate (\%) $\uparrow$}} &
\multicolumn{5}{c}{\textbf{SD1.5 Metrics $\uparrow$}} \\
\cmidrule(lr){3-8}\cmidrule(lr){9-13}
 & &
PS & HPS & CLIP & IM & AES & Avg &
PS & HPS & CLIP & IM & AES \\
\midrule
\multirow{6}{*}{Pick-a-Pic} & Original & -- & -- & -- & -- & -- & -- & 20.68 & 26.74 & 33.23 & 0.0495 & 5.5060 \\
 & Online-SFT & 40.44 & 46.60 & 54.96 & 53.24 & 59.28 & 50.90 & 20.51 & 26.64 & 33.97 & 0.1555 & 5.5942 \\
 & Online-DPO & 77.40 & 82.48 & \underline{59.96} & \underline{81.00} & 73.36 & 74.84 & 21.26 & 27.79 & \textbf{34.62} & \underline{0.6861} & 5.7498 \\
 & Seppo & \textbf{82.68} & \underline{84.52} & 55.04 & 75.32 & 80.56 & \underline{75.62} & 21.40 & \underline{27.80} & 33.93 & 0.5381 & 5.8106 \\
 & SPIN & 81.32 & 77.36 & 44.68 & 69.88 & \textbf{85.36} & 71.72 & \underline{21.43} & 27.65 & 32.54 & 0.4461 & \textbf{5.9669} \\
 & \cellcolor{yellow!20}Diff.-NPO& \cellcolor{yellow!20} \underline{81.60}& \cellcolor{yellow!20} \textbf{84.64}& \cellcolor{yellow!20} \textbf{60.32}& \cellcolor{yellow!20} \textbf{85.12}& \cellcolor{yellow!20} \underline{80.76}& \cellcolor{yellow!20} \textbf{78.49}& \cellcolor{yellow!20} \textbf{21.48}& \cellcolor{yellow!20} \textbf{28.02}& \cellcolor{yellow!20} \underline{34.61}& \cellcolor{yellow!20} \textbf{0.8896}& \cellcolor{yellow!20} \underline{5.8773} \\
\hline
\multirow{6}{*}{\shortstack{Parti-\\Prompts}} & Original & -- & -- & -- & -- & -- & -- & 21.53 & 27.44 & 33.15 & 0.1926 & 5.3592 \\
 & Online-SFT & 38.88 & 45.24 & \underline{55.10} & 54.18 & 59.83 & 50.65 & 21.35 & 27.37 & 33.84 & 0.3366 & 5.4508 \\
 & Online-DPO & 71.43 & \underline{78.42} & \textbf{56.38} & \underline{74.73} & 74.73 & \underline{71.14} & 21.95 & \underline{28.31} & \textbf{34.09} & \underline{0.6699} & 5.5266 \\
 & Seppo & \underline{73.97} & 77.59 & 50.44 & 68.58 & 75.56 & 69.23 & 21.97 & 28.24 & 33.43 & 0.5143 & 5.5904 \\
 & SPIN & 72.73 & 72.34 & 39.01 & 62.34 & \textbf{81.92} & 65.67 & \underline{21.98} & 28.15 & 31.98 & 0.4014 & \textbf{5.7324} \\
 & \cellcolor{yellow!20}Diff.-NPO& \cellcolor{yellow!20} \textbf{74.76}& \cellcolor{yellow!20} \textbf{79.70}& \cellcolor{yellow!20} 54.74& \cellcolor{yellow!20} \textbf{77.04}& \cellcolor{yellow!20} \underline{76.96}& \cellcolor{yellow!20} \textbf{72.64}& \cellcolor{yellow!20} \textbf{22.07}& \cellcolor{yellow!20} \textbf{28.49}& \cellcolor{yellow!20} \underline{34.07}& \cellcolor{yellow!20} \textbf{0.7966}& \cellcolor{yellow!20} \underline{5.6231} \\
\hline
\multirow{6}{*}{HPSV2} & Original & -- & -- & -- & -- & -- & -- & 21.05 & 27.08 & 35.43 & 0.0727 & 5.5725 \\
 & Online-SFT & 40.25 & 48.70 & 50.25 & 55.15 & 57.95 & 50.46 & 21.27 & 27.06 & 35.53 & 0.2053 & 5.6453 \\
 & Online-DPO & 80.70 & 85.10 & \underline{59.20} & \underline{80.35} & 73.80 & 75.83 & 21.75 & 28.21 & \underline{36.51} & \underline{0.7092} & 5.8203 \\
 & Seppo & \textbf{83.75} & \underline{85.75} & 54.10 & 77.80 & \underline{81.80} & \underline{76.64} & 21.88 & \underline{28.27} & 35.99 & 0.6121 & \underline{5.9132} \\
 & SPIN & 81.70 & 80.65 & 44.65 & 71.10 & \textbf{87.25} & 73.07 & \underline{21.92} & 28.20 & 34.69 & 0.5027 & \textbf{6.0577} \\
 & \cellcolor{yellow!20}Diff.-NPO& \cellcolor{yellow!20} \underline{83.15}& \cellcolor{yellow!20} \textbf{87.75}& \cellcolor{yellow!20} \textbf{60.05}& \cellcolor{yellow!20} \textbf{83.90}& \cellcolor{yellow!20} 79.45& \cellcolor{yellow!20} \textbf{78.86}& \cellcolor{yellow!20} \textbf{21.96}& \cellcolor{yellow!20} \textbf{28.53}& \cellcolor{yellow!20} \textbf{36.65}& \cellcolor{yellow!20} \textbf{0.8761}& \cellcolor{yellow!20} 5.8957 \\
\bottomrule
\end{tabular}%
\end{table}

%% file: table/sdxl_result.tex
\begin{table}[h!]
\centering
\small
\setlength{\tabcolsep}{1.5pt}
\renewcommand{\arraystretch}{1.0}
\caption{SDXL win-rate and automatic metric results for \ourmech compared with other baselines. Win rates are vs.~original SDXL. Yellow rows indicate our method. Best is in \textbf{bold}; second-best is \underline{underlined}.}
\label{tab:sdxl-result}
\vspace{-0.75em}
{%
\begin{tabular}{l l *{6}{C{2.75em}} *{3}{C{2.75em}} *{2}{C{3.45em}} }
\toprule 
\multirow{2}{*}{\textbf{Dataset}} & \multirow{2}{*}{\textbf{Method}} &
\multicolumn{6}{c}{\textbf{SDXL Winrate (\%) $\uparrow$}} &
\multicolumn{5}{c}{\textbf{SDXL Metrics $\uparrow$}} \\
\cmidrule(lr){3-8}\cmidrule(lr){9-13}
 & &
PS & HPS & CLIP & IM & AES & Avg &
PS & HPS & CLIP & IM & AES \\
\midrule
\multirow{4}{*}{Pick-a-Pic} & Original & -- & -- & -- & -- & -- & -- & 22.06 & 22.71 & 36.38 & 0.5562 & 6.0735 \\
 & Online-SFT & 28.84 & 38.00 & \textbf{57.84} & 53.36 & 37.40 & 43.09 & 21.72 & 27.46 & \textbf{37.15} & 0.6539 & 5.9530 \\
 & Online-DPO & \underline{82.08} & \underline{81.92} & \underline{56.16} & \underline{85.32} & \underline{65.64} & \underline{74.22} & \underline{22.71} & \underline{28.57} & \underline{37.05} & \underline{1.1247} & \underline{6.2686} \\
 & \cellcolor{yellow!20} Diff.-NPO& \cellcolor{yellow!20} \textbf{82.64}& \cellcolor{yellow!20} \textbf{86.12}& \cellcolor{yellow!20} 55.12& \cellcolor{yellow!20} \textbf{89.56}& \cellcolor{yellow!20} \textbf{74.96}& \cellcolor{yellow!20} \textbf{77.68}& \cellcolor{yellow!20} \textbf{22.78}& \cellcolor{yellow!20} \textbf{28.85}& \cellcolor{yellow!20} 37.01& \cellcolor{yellow!20} \textbf{1.2599}& \cellcolor{yellow!20} \textbf{6.4294} \\
\hline
\multirow{4}{*}{\shortstack{Parti-\\Prompts}} & Original & -- & -- & -- & -- & -- & -- & 22.51 & 28.00 & 35.23 & 0.6003 & 5.7438 \\
 & Online-SFT & 30.21 & 36.02 & \textbf{54.42} & 51.38 & 44.13 & 43.23 & 22.21 & 27.68 & \textbf{35.86} & 0.6976 & 5.6925 \\
 & Online-DPO & \underline{78.14} & \underline{77.98} & \underline{47.32} & \underline{79.88} & \underline{76.74} & \underline{72.01} & \underline{23.07} & \underline{28.85} & 35.26 & \underline{1.0427} & \underline{6.0486} \\
 & \cellcolor{yellow!20} Diff.-NPO& \cellcolor{yellow!20} \textbf{82.66}& \cellcolor{yellow!20} \textbf{84.88}& \cellcolor{yellow!20} 45.81& \cellcolor{yellow!20} \textbf{86.73}& \cellcolor{yellow!20} \textbf{85.04}& \cellcolor{yellow!20} \textbf{77.02}& \cellcolor{yellow!20} \textbf{23.26}& \cellcolor{yellow!20} \textbf{29.18}& \cellcolor{yellow!20} \underline{35.34}& \cellcolor{yellow!20} \textbf{1.2505}& \cellcolor{yellow!20} \textbf{6.2496} \\
\hline
\multirow{4}{*}{HPSV2} & Original & -- & -- & -- & -- & -- & -- & 22.75 & 28.18 & 38.64 & 0.7526 & 6.1155 \\
 & Online-SFT & 21.85 & 34.70 & 48.20 & 47.25 & 35.75 & 37.55 & 22.25 & 27.84 & 38.62 & 0.7715 & 6.0088 \\
 & Online-DPO & \textbf{84.80} & \underline{84.75} & \textbf{52.65} & \underline{83.35} & \underline{70.55} & \underline{75.22} & \underline{23.46} & \underline{29.17} & \textbf{38.92} & \underline{1.2205} & \underline{6.3367} \\
 & \cellcolor{yellow!20} Diff.-NPO& \cellcolor{yellow!20} \underline{83.65}& \cellcolor{yellow!20} \textbf{88.60}& \cellcolor{yellow!20} \underline{50.69}& \cellcolor{yellow!20} \textbf{88.60}& \cellcolor{yellow!20} \textbf{75.10}& \cellcolor{yellow!20} \textbf{77.33}& \cellcolor{yellow!20} \textbf{23.49}& \cellcolor{yellow!20} \textbf{29.42}& \cellcolor{yellow!20} \underline{38.75}& \cellcolor{yellow!20} \textbf{1.3319}& \cellcolor{yellow!20} \textbf{6.4344} \\
\bottomrule
\end{tabular}%
}\vspace{-0.5em}
\end{table}

%% file: table/ablation_ratio.tex
\centering
\small
\setlength{\tabcolsep}{2pt}
\renewcommand{\arraystretch}{1.1}
\begin{tabular}{l | c c c c c}
\hline
\textbf{$\tau/\eta$} &
PS $\uparrow$ & HPS $\uparrow$ & CLIP $\uparrow$ & IM $\uparrow$ & AES $\uparrow$ \\
\hline
$1$ & 21.26 & 27.79 & 34.62 & 0.6861 & 5.7498 \\
$8/9$ & 21.40 & 27.92 & 34.59 & 0.8001 & 5.7854 \\
$1/2$ & 21.48 & 28.02 & 34.61 & 0.8896 & 5.8773 \\
$1/3$ & 21.34 & 27.76 & 33.97 & 0.7817 & 5.7916 \\
$1/9$ & 21.29 & 27.89 & 34.12 & 0.8098 & 5.9041 \\
$0$ & 20.87 & 27.41 & 33.78 & 0.4875 & 5.7541 \\
\hline
\end{tabular}

%% file: table/ablation_square_loss.tex
\centering
\small
\renewcommand{\arraystretch}{0.97}
\setlength{\tabcolsep}{2pt}
\begin{tabular}{lccccc}
\toprule
\multicolumn{6}{c}{\textbf{Win Rate (\%) on Pick-a-Pic}} \\
\midrule
Obj. & PS $\uparrow$ & HPS $\uparrow$ & CLIP $\uparrow$ & IM $\uparrow$ & AES $\uparrow$ \\
\midrule
SD & 60.80 & 74.60 & 54.80 & 70.20 & 71.60 \\
PL & \textbf{81.60} & \textbf{84.64} & \textbf{60.32} & \textbf{85.12} & \textbf{80.76} \\
\bottomrule
\end{tabular}

%% file: appendix.tex
\section{\ourmech Pseudo-code}\label{app:algo}

The Pseudo-code of \ourmech is shown in Algorithm~\ref{algo:npo}.

\input{table/algo}

\section{Related Work}\label{app:relatedwork}

\input{2_related_work.tex}

\section{Mathematical Analysis}\label{app:math-analysis}

\subsection{Derivation of the OMD Update for Nash Policy}

Given the preference probability $\mathbb{P}(\rvx_0\succ \rvx_0'\mid \rvc)$, define the loss at step $s$ as
\begin{align*}
\ell_s(p)
&:=
-\mathbb{E}_{\rvc\sim d_0}\,\mathbb{E}_{\rvx_0\sim p(\rvx_0 \mid \rvc),\,\rvx_0'\sim p_s(\rvx_0\mid \rvc)}\big[\mathbb{P}(\rvx_0\succ \rvx_0'\mid \rvc)\big]
+\tau\,\mathrm{KL}\big(p(\rvx_0\mid \rvc)\,\|\,p_{\mathrm{ref}}(\rvx_0\mid \rvc)\big) \\
& \leq 
-\mathbb{E}_{\rvc\sim d_0}\,\mathbb{E}_{\rvx_0\sim p(\rvx_0 \mid \rvc),\,\rvx_0'\sim p_s(\rvx_0\mid \rvc)}\big[\mathbb{P}(\rvx_0\succ \rvx_0'\mid \rvc)\big]
+\tau\,\mathrm{KL}\big(p(\rvx_{0:T}\mid \rvc)\,\|\,p_{\mathrm{ref}}(\rvx_{0:T}\mid \rvc)\big) \\
& =
-\mathbb{E}_{\rvc\sim d_0}\,\mathbb{E}_{\rvx_{0:T}\sim p(\rvx_{0:T} \mid \rvc),\,\rvx_{0:T}'\sim p_s(\rvx_{0:T}\mid \rvc)}\big[\mathbb{P}(\rvx_{0:T} \succ \rvx_{0:T}'\mid \rvc)\big] \\
& \hspace{20em} +\tau\,\mathrm{KL}\big(p(\rvx_{0:T}\mid \rvc)\,\|\,p_{\mathrm{ref}}(\rvx_{0:T}\mid \rvc)\big), 
\end{align*}
Since $\mathrm{KL}\big(p(\rvx_0\mid \rvc)\,\|\,p_{\mathrm{ref}}(\rvx_0\mid \rvc)\big)$ is not directly tractable, we work on its upper bound $\mathrm{KL}\big(p(\rvx_{0:T}\mid \rvc)\,\|\,p_{\mathrm{ref}}(\rvx_{0:T}\mid \rvc)\big)$, which lower-bounds the objective. Starting from the entropy-regularized online mirror descent update,
\[
p_{s+1}
=
\arg\min_{p\in \mathcal{P}}
\Big\langle \nabla \ell_s(p_s),p\Big\rangle
+
\eta\,\mathrm{KL}(p\|p_s),
\]
The objective can be written (for a fixed prompt $\rvc$) as
\begin{align*}
&\min_{p}
\sum_{\rvx_{0:T}}p(\rvx_{0:T}\mid \rvc)\,\nabla_{p_s(\rvx_{0:T})}\ell_s(p_s)
+
\eta\sum_{\rvx_{0:T}}p(\rvx_{0:T}\mid \rvc)\log\frac{p(\rvx_{0:T}\mid \rvc)}{p_s(\rvx_{0:T}\mid \rvc)}, \\
&\text{s.t. } \sum_{\rvx_{0:T}}p(\rvx_{0:T}\mid \rvc)=1.
\end{align*}
Introducing a Lagrange multiplier $\lambda$ for the normalization constraint, the Lagrangian is
\begin{align*}
\mathcal L(p,\lambda)
=
\sum_{\rvx_{0:T}}p(\rvx_{0:T}\mid \rvc)\,\nabla_{p_s(\rvx_{0:T})}\ell_s(p_s)
+
\eta\sum_{\rvx_{0:T}} & p(\rvx_{0:T}\mid \rvc)\log\frac{p(\rvx_{0:T}\mid \rvc)}{p_s(\rvx_{0:T}\mid \rvc)} \\
& +
\lambda\Big(\sum_{\rvx_{0:T}}p(\rvx_{0:T}\mid \rvc)-1\Big).
\end{align*}
When the derivative of $\mathcal L(p,\lambda)$ with respect to $p(\rvx_{0:T}\mid \rvc)$ goes to zero, we get
\[
\log\tfrac{p_{s+1}(\rvx_{0:T}\mid \rvc)}{p_s(\rvx_{0:T}\mid \rvc)}
=
-\tfrac{1}{\eta}\nabla_{p_s(\rvx_{0:T})}\ell_s(p_s)-1-\tfrac{\lambda}{\eta},
\]
after exponentiating the left and right sides:
\begin{align*}
p_{s+1}(\rvx_{0:T}\mid \rvc)
&\propto
p_s(\rvx_{0:T}\mid \rvc)\exp\Big(-\tfrac{1}{\eta}\nabla_{p_s(\rvx_{0:T})}\ell_s(p_s)\Big), \\
&\propto
p_s(\rvx_{0:T}\mid \rvc)
\exp\Big(\tfrac{1}{\eta}\,\mathbb{P}(\rvx_{0:T}\succ p_s\mid \rvc)\Big)
\Big(\tfrac{p_{\mathrm{ref}}(\rvx_{0:T}\mid \rvc)}{p_s(\rvx_{0:T}\mid \rvc)}\Big)^{\tfrac{\tau}{\eta}},
\end{align*}

which can be collected as
\[
p_{s+1}(\rvx_{0:T}\mid \rvc)
\propto
\exp\Big(\tfrac{1}{\eta}\,\mathbb{P}(\rvx_{0:T}\succ p_s\mid \rvc)\Big)
\,p_{\mathrm{ref}}(\rvx_{0:T}\mid \rvc)^{\tfrac{\tau}{\eta}}
\,p_s(\rvx_{0:T}\mid \rvc)^{1-\tfrac{\tau}{\eta}}.
\]

\subsection{Derivation of Final Objective (Equation~\ref{eq:obj-final})}\label{app:per-step-obj}

To instantiate the full-chain policy objective in Equation~\ref{eq:obj-chain} for diffusion models, we identify the generic policy notation with the corresponding diffusion reverse process. Specifically, the current policy $\pi$ is parameterized by the trainable diffusion model $p_\theta$, the previous policy $\pi_s$ is denoted by $p_{\mathrm{prev}}$, and the reference policy $\pi_{\mathrm{ref}}$ is denoted by $p_{\mathrm{ref}}$. Under the Markov structure of the diffusion reverse process, the full-chain likelihood decomposes as
\[
p_\theta(\rvx_{0:T}\mid \rvc)
=
p(\rvx_T)\prod_{t=1}^T p_\theta(\rvx_{t-1}\mid \rvx_t,\rvc),
\]
and similarly for $p_{\mathrm{ref}}$ and $p_{\mathrm{prev}}$. Therefore, the log-ratio over full trajectories in Equation~\ref{eq:obj-chain} can be decomposed into a sum of per-step reverse-transition log-ratios. Following the standard diffusion-DPO approximation, we sample a timestep $t\sim \mathcal{U}(0,T)$, replace the intractable reverse-chain sampling with the forward noising process $q(\rvx_t\mid \rvx_0)$, and optimize the resulting single-timestep objective with the factor $T$ accounting for uniform timestep sampling.

We start with Equation~\ref{eq:obj-chain}, the objective on reverse decomposition for current, reference, and previous policies,
\begin{equation*}
\begin{aligned}
& \mathbb{E}_{\rvx^+_{0:T},\rvx^-_{0:T}\sim\pi_t(\cdot)}\Big[
  -\log \sigma\Big(
    \frac{\tau \beta}{\eta}\Big(
      \log\frac{\pi(\rvx^+_{0:T})}{\pi(\rvx^-_{0:T})} 
      -\log\frac{\pi_{\mathrm{ref}}(\rvx^+_{0:T})}{\pi_{\mathrm{ref}}(\rvx^-_{0:T})}
    \Big) \\
   &\hspace{18em} + \frac{(\eta-\tau)\beta}{\eta}\Big(
      \log\frac{\pi(\rvx^+_{0:T})}{\pi(\rvx^-_{0:T})}
      -\log\frac{\pi_t(\rvx^+_{0:T})}{\pi_t(\rvx^-_{0:T})}
    \Big)
  \Big)
\Big]\\
=& \mathbb{E}_{\rvx^+_{1:T}\sim q(\rvx^+_{1:T} \mid \rvx^+_0),\rvx^-_{1:T}\sim q(\rvx^-_{1:T} \mid \rvx^-_0)}\Big[
  -\log \sigma\Big(
    \frac{\tau \beta}{\eta}\Big(
       \log\frac{p_\theta(\rvx^+_{0:T})}{p_\theta(\rvx^-_{0:T})}
      -\log\frac{p_{\mathrm{ref}}(\rvx^+_{0:T})}{p_{\mathrm{ref}}(\rvx^-_{0:T})}
    \Big) \\
    &\hspace{16em}
    + \frac{(\eta-\tau)\beta}{\eta}\Big(
      \log\frac{p_\theta(\rvx^+_{0:T})}{p_\theta(\rvx^-_{0:T})}
      -\log\frac{p_\mathrm{prev}(\rvx^+_{0:T})}{p_\mathrm{prev}(\rvx^-_{0:T})}
    \Big)
  \Big)
\Big],
\end{aligned}
\end{equation*}
where we use $q(\rvx_{1:T} \mid \rvx_0)$ to approximate $p_{\mathrm{prev}}(\rvx_{1:T} \mid \rvx_0)$ for efficient training. The objective hence becomes,
\begin{equation*}
    \begin{aligned}
        L = & \mathbb{E}_{\rvx^+_{1:T}\sim q(\rvx^+_{1:T} \mid \rvx^+_0),\rvx^-_{1:T}\sim q(\rvx^-_{1:T} \mid \rvx^-_0)}\Big[
  -\log \sigma\Big(
    \frac{\tau \beta}{\eta}\Big(
       \log\frac{p_\theta(\rvx^+_{0:T})}{p_\theta(\rvx^-_{0:T})}
      -\log\frac{p_{\mathrm{ref}}(\rvx^+_{0:T})}{p_{\mathrm{ref}}(\rvx^-_{0:T})}
    \Big) \\
    &
    + \frac{(\eta-\tau)\beta}{\eta}\Big(
      \log\frac{p_\theta(\rvx^+_{0:T})}{p_\theta(\rvx^-_{0:T})}
      -\log\frac{p_\mathrm{prev}(\rvx^+_{0:T})}{p_\mathrm{prev}(\rvx^-_{0:T})}
    \Big)
  \Big)
\Big] \\
 = &
\mathbb{E}_{\rvx^+_{1:T}\sim q(\rvx^+_{1:T} \mid \rvx^+_0),\rvx^-_{1:T}\sim q(\rvx^-_{1:T} \mid \rvx^-_0)}\Big[
  -\log \sigma\Big(\sum_{t=0}^{T-1}
    \frac{\tau \beta}{\eta}\Big(
       \log\frac{p_\theta(\rvx^+_{t-1}|\rvx^+_{t})}{p_\theta(\rvx^-_{t-1}|\rvx^-_{t})}
      -\log\frac{p_{\mathrm{ref}}(\rvx^+_{t-1}|\rvx^+_{t})}{p_{\mathrm{ref}}(\rvx^-_{t-1}|\rvx^-_{t})}
    \Big) \\
    &
    + \frac{(\eta-\tau)\beta}{\eta}\Big( 
      \log\frac{p_\theta(\rvx^+_{t-1}|\rvx^+_{t})}{p_\theta(\rvx^-_{t-1}|\rvx^-_{t})}
      -\log\frac{p_\mathrm{prev}(\rvx^+_{t-1}|\rvx^+_{t})}{p_\mathrm{prev}(\rvx^-_{t-1}|\rvx^-_{t})}
    \Big)
  \Big)
\Big] \\
= &
\mathbb{E}_{\rvx^+_{1:T}\sim q(\rvx^+_{1:T} \mid \rvx^+_0),\rvx^-_{1:T}\sim q(\rvx^-_{1:T} \mid \rvx^-_0)}\Big[
  -\log \sigma\Big(T \mathbb{E}_t\Big[
    \frac{\tau \beta}{\eta}\Big(
       \log\frac{p_\theta(\rvx^+_{t-1}|\rvx^+_{t})}{p_\theta(\rvx^-_{t-1}|\rvx^-_{t})}
      -\log\frac{p_{\mathrm{ref}}(\rvx^+_{t-1}|\rvx^+_{t})}{p_{\mathrm{ref}}(\rvx^-_{t-1}|\rvx^-_{t})}
    \Big) \\
    &
    + \frac{(\eta-\tau)\beta}{\eta}\Big( 
      \log\frac{p_\theta(\rvx^+_{t-1}|\rvx^+_{t})}{p_\theta(\rvx^-_{t-1}|\rvx^-_{t})}
      -\log\frac{p_\mathrm{prev}(\rvx^+_{t-1}|\rvx^+_{t})}{p_\mathrm{prev}(\rvx^-_{t-1}|\rvx^-_{t})}
    \Big) \Big]
  \Big)
\Big] \\
= &
\mathbb{E}_{\rvx^+_{t, t-1}\sim q(\rvx^+_{t, t-1} \mid \rvx^+_0),\rvx^-_{t, t-1}\sim q(\rvx^-_{t, t-1} \mid \rvx^-_0)}\Big[
  -\log \sigma\Big(T \mathbb{E}_t\Big[
    \frac{\tau \beta}{\eta}\Big(
       \log\frac{p_\theta(\rvx^+_{t-1}|\rvx^+_{t})}{p_\theta(\rvx^-_{t-1}|\rvx^-_{t})} \\
    &
    - \log\frac{p_{\mathrm{ref}}(\rvx^+_{t-1}|  \rvx^+_{t})}{p_{\mathrm{ref}}(\rvx^-_{t-1}| \rvx^-_{t})}
    \Big) + \frac{(\eta-\tau)\beta}{\eta}\Big( 
      \log\frac{p_\theta(\rvx^+_{t-1}|\rvx^+_{t})}{p_\theta(\rvx^-_{t-1}|\rvx^-_{t})}
      -\log\frac{p_\mathrm{prev}(\rvx^+_{t-1}|\rvx^+_{t})}{p_\mathrm{prev}(\rvx^-_{t-1}|\rvx^-_{t})}
    \Big) \Big]
  \Big)
\Big]. 
    \end{aligned}
\end{equation*}
By applying Jensen's inequality, we can lower bound the objective as
\begin{equation*}
    \begin{aligned}
        L &\leq \mathbb{E}_{\substack{t,\, \rvx^+_{t}\sim q(\rvx^+_{t} \mid \rvx^+_0),\\
        \rvx^-_{t}\sim q(\rvx^-_{t} \mid \rvx^-_0)}}\Big[
        -\log \sigma\Big(T \mathbb{E}_{\substack{\rvx^+_{t-1}\sim q(\rvx^+_{t-1} \mid \rvx^+_{t}, \rvx^+_0),\\
        \rvx^-_{t-1}\sim q(\rvx^-_{t-1} \mid \rvx^-_{t}, \rvx^-_0)}} \Big[
            \frac{\tau \beta}{\eta}\Big(
                \log\frac{p_\theta(\rvx^+_{t-1}|\rvx^+_{t})}{p_\theta(\rvx^-_{t-1}|\rvx^-_{t})} \\
              &  -\log\frac{p_{\mathrm{ref}}(\rvx^+_{t-1}|\rvx^+_{t})}{p_{\mathrm{ref}}(\rvx^-_{t-1}|\rvx^-_{t})}\Big)
             + \frac{(\eta-\tau)\beta}{\eta}\Big(
                \log\frac{p_\theta(\rvx^+_{t-1}|\rvx^+_{t})}{p_\theta(\rvx^-_{t-1}|\rvx^-_{t})}
                -\log\frac{p_\mathrm{prev}(\rvx^+_{t-1}|\rvx^+_{t})}{p_\mathrm{prev}(\rvx^-_{t-1}|\rvx^-_{t})}
            \Big)
        \Big]\Big)\Big] \\
        &= \mathbb{E}_{\substack{t,\, \rvx^+_{t}\sim q(\rvx^+_{t} \mid \rvx^+_0),\\
        \rvx^-_{t}\sim q(\rvx^-_{t} \mid \rvx^-_0)}}\Big[
        -\log \sigma\Big(-T \Big[
            \frac{\tau \beta}{\eta}\Big(
                \mathbb{D}_{\mathrm{KL}}(q(\rvx_{t-1}^+ | \rvx_{0,t}^+) \|\, p_\theta(\rvx^+_{t-1}|\rvx^+_{t})) \\
                & \qquad\qquad - \mathbb{D}_{\mathrm{KL}}(q(\rvx_{t-1}^- | \rvx_{0,t}^-) \|\, p_\theta(\rvx^-_{t-1}|\rvx^-_{t}))
            \Big. \\
            &\qquad\qquad\Big.
                - \mathbb{D}_{\mathrm{KL}}(q(\rvx_{t-1}^+ | \rvx_{0,t}^+) \|\, p_{\mathrm{ref}}(\rvx^+_{t-1}|\rvx^+_{t}))
                + \mathbb{D}_{\mathrm{KL}}(q(\rvx_{t-1}^- | \rvx_{0,t}^-) \|\, p_{\mathrm{ref}}(\rvx^-_{t-1}|\rvx^-_{t}))
            \Big) \\
            &\qquad + \frac{(\eta-\tau)\beta}{\eta}\Big(
                \mathbb{D}_{\mathrm{KL}}(q(\rvx_{t-1}^+ | \rvx_{0,t}^+) \|\, p_\theta(\rvx^+_{t-1}|\rvx^+_{t}))
                - \mathbb{D}_{\mathrm{KL}}(q(\rvx_{t-1}^- | \rvx_{0,t}^-) \|\, p_\theta(\rvx^-_{t-1}|\rvx^-_{t}))
            \Big. \\
            &\qquad\Big.
                - \mathbb{D}_{\mathrm{KL}}(q(\rvx_{t-1}^+ | \rvx_{0,t}^+) \|\, p_{\mathrm{prev}}(\rvx^+_{t-1}|\rvx^+_{t}))
                + \mathbb{D}_{\mathrm{KL}}(q(\rvx_{t-1}^- | \rvx_{0,t}^-) \|\, p_{\mathrm{prev}}(\rvx^-_{t-1}|\rvx^-_{t}))
            \Big)
        \Big]\Big)\Big].
    \end{aligned}
\end{equation*}
Since the distribution of reverse step $t$ of diffusion models $p_\theta(\rvx_0)$ is
\begin{align*}
    p_\theta(\rvx_{t-1}|\rvx_t)
    &= \mathcal{N}(\rvx_{t-1};\sqrt{\frac{\alpha_{t-1}}{\alpha_{t}}}\left(\rvx_{t} - \frac{\beta_{t}}{\sqrt{1-\bar{\alpha}_{t}}} \epsilon(\rvx_{t})\right), \sigma_{t}^2 I),
\end{align*}
we can parameterize the loss as
\begin{equation*}
    \begin{aligned}
        L &\leq \mathbb{E}_{\substack{t,\, \rvx^+_{t}\sim q(\rvx^+_{t} \mid \rvx^+_0),\\
        \rvx^-_{t}\sim q(\rvx^-_{t} \mid \rvx^-_0)}}\Big[
        -\log \sigma\Big(-T \Big[
            \frac{\tau \beta}{\eta}\Big(
                \lVert\epsilon^+ - \epsilon_\theta(\rvx_t^+, t)\rVert^2
                - \lVert\epsilon^- - \epsilon_\theta(\rvx_t^-, t)\rVert^2
            \Big. \\
            &\qquad\qquad\Big.
                - \lVert\epsilon^+ - \epsilon_{\mathrm{ref}}(\rvx_t^+, t)\rVert^2
                + \lVert\epsilon^- - \epsilon_{\mathrm{ref}}(\rvx_t^-, t)\rVert^2
            \Big) \\
            &\qquad + \frac{(\eta-\tau)\beta}{\eta}\Big(
                \lVert\epsilon^+ - \epsilon_\theta(\rvx_t^+, t)\rVert^2
                - \lVert\epsilon^- - \epsilon_\theta(\rvx_t^-, t)\rVert^2
            \Big. \\
            &\qquad\qquad\Big.
                - \lVert\epsilon^+ - \epsilon_{\mathrm{prev}}(\rvx_t^+, t)\rVert^2
                + \lVert\epsilon^- - \epsilon_{\mathrm{prev}}(\rvx_t^-, t)\rVert^2
            \Big)
        \Big]\Big)\Big]
    \end{aligned}
\end{equation*}

\section{Additional Experimental Details}\label{app:exp-details}

\subsection{Datasets}
\paragraph{Pick-a-Pic.}
Pick-a-Pic is a large-scale public dataset of human preferences for text-to-image generation~\cite{kirstain2023pickapicopendatasetuser} with  over 500,000 examples. It is collected through a web interface in which users provide prompts, compare multiple generated images, and indicate which image they prefer, or whether the comparison results in a tie. Each example therefore consists of a prompt, two generated images, and a preference label. We use Pick-a-Pic v1 for training and its held-out test split for evaluation.

\paragraph{PartiPrompts.}
PartiPrompts is a curated benchmark designed to evaluate the compositional and semantic capabilities of text-to-image models~\cite{yu2022scalingautoregressivemodelscontentrich}. It contains prompts covering a broad range of visual concepts, including objects, attributes, styles, and complex relational descriptions. Owing to this diversity, PartiPrompts has become a standard benchmark for assessing prompt following, compositional generation, and overall image quality. In our experiments, we use PartiPrompts as a standardized evaluation suite for text-to-image generation.

\paragraph{HPS v2.}
HPS v2 is a human-preference benchmark built from over
25,000 prompts and 98,000 images generated by Stable Diffusion, accompanied by 25,205 human
preference annotations collected from the Stable Foundation Discord community~\cite{wu2023humanpreferencescorev2}. For each prompt, annotators compare multiple candidate images and indicate the one they prefer. These annotations are then used to train a CLIP-based preference model, which defines the Human Preference Score (HPS). We use the HPS v2 test set to evaluate how well our aligned diffusion models reflect human preferences.

\subsection{Baselines}
We compare \ourmech against several representative baselines. The first baseline is the original pretrained diffusion backbone, which provides the starting point for alignment on both SD1.5 and SDXL. We then consider online supervised fine-tuning (SFT) on preferred images, including both offline and online variants. We also compare with online-Diff-DPO in the main tables. In addition, we include two representative self-play preference optimization methods, SPIN~\citep{SPIN} and SEPPO~\citep{SEPPO}, with detailed introductions in paragraphs below. Together, these baselines allow us to compare our method with reference-anchored preference optimization, pure self-play style updates, and simple supervised adaptation.

\paragraph{SPIN.}
SPIN is a self-play fine-tuning method for text-to-image diffusion models. 
Instead of relying on a fixed offline preference dataset, SPIN iteratively improves the model by comparing samples generated by the current model with samples from a previous or reference model. 
At each iteration, the model is encouraged to assign higher likelihood to its own improved generations than to weaker generations produced by an earlier policy. 
This creates a self-play learning dynamic, where the model continuously learns to outperform its past versions. 
In our comparison, SPIN represents a pure self-play style baseline: it uses online model-generated samples to construct preference pairs and update the diffusion model, but does not explicitly combine both reference-policy anchoring and previous-policy regularization in the same way as \ourmech.

\paragraph{SEPPO.}
SEPPO is a semi-policy preference optimization method for diffusion alignment. 
Its main idea is to bridge offline preference optimization and online self-play by using samples generated from both the current policy and an auxiliary policy, such as a reference or previous policy, to construct preference pairs. 
Compared with standard Diffusion-DPO, which is mainly anchored to a fixed reference model, SEPPO introduces policy-dependent samples so that the training distribution can better reflect the model's current generation behavior. 
Compared with pure self-play methods, however, SEPPO still maintains a semi-policy structure rather than fully optimizing the current policy through an explicit Nash-style online mirror descent objective. 

\subsection{Implementation Details}
We adopt AdamW for training SD1.5 and Adafactor for training SDXL. The learning rate is set to $1 \times 10^{-8}$ with linear warmup and is scaled by the effective batch size. The global batch size is 2048: for pairwise methods, the effective batch contains 2048 pairs. For SD1.5 and SDXL, we set $\beta = 5000$ for Online-Diffusion-DPO and Diff.-NPO. For Diff.-NPO, we fix the ratio $\tau/\eta=1/2$ in all experiments unless the ablation study. For SEPPO and SPIN, we directly evaluate the official checkpoints released by their authors. All experiments are conducted on eight NVIDIA A100 80GB GPUs.

\section{Additional Ablation Results}

\subsection{Ablation on $\tau/\eta$ Regularization}\label{app:reg-ablation}

\begin{figure}[h]
  \centering
  \includegraphics[width=\textwidth]{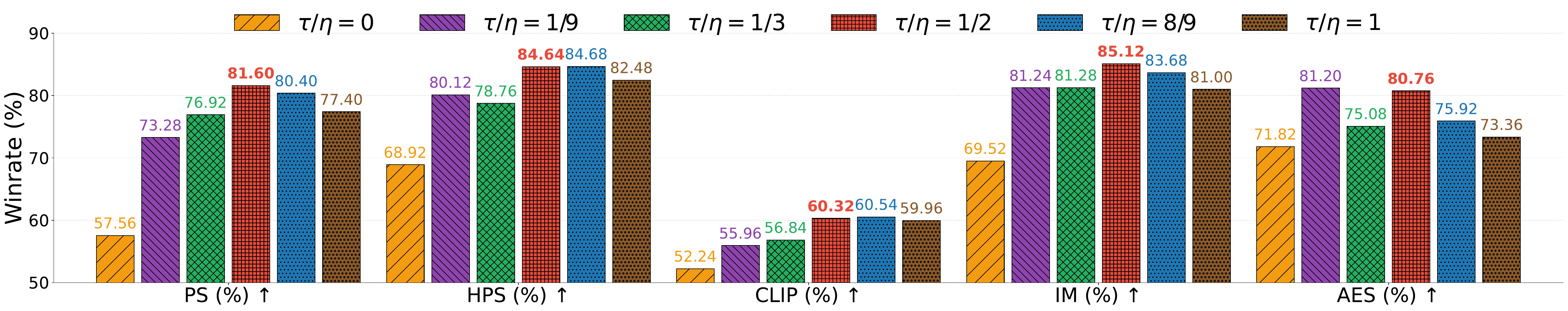}
  \caption{Full win-rate comparison for the ablation study of different $\tau/\eta$ values on Pick-a-Pic for SD1.5. We report win rates across all five evaluation metrics: PickScore, HPSV2, CLIP score, ImageReward, and Aesthetic Score.}
  \label{fig:ablation-all-5}
\end{figure}

Figure~\ref{fig:ablation-all-5} presents the complete win-rate comparison for different $\tau/\eta$ values across all five evaluation metrics. Consistent with the main-text results, the performance generally improves from the pure self-play setting and then decreases as the objective approaches the DPO-style setting. This trend suggests that intermediate regularization ratios better balance the stabilizing effect of the reference policy and the online self-improvement signal from the previous policy.

\subsection{Comparison with Offline SFT and DPO}

We further compare \ourmech with offline variants of SFT and DPO in Table~\ref{tab:offline}. 
Offline SFT substantially outperforms online SFT, mainly because the preferred images in the offline Pick-a-Pic dataset are generally of higher quality than the images generated online with a small number of inference steps. 
Nevertheless, offline SFT still underperforms preference-optimization methods, suggesting that training only on preferred images cannot fully exploit the relative preference information contained in positive--negative pairs. 
We also observe that offline DPO is weaker than online DPO in this setting, indicating that online model-generated preference pairs better match the current policy distribution. 
Overall, Diff.-NPO achieves the best win rates across all five metrics, showing that the proposed online Nash-style objective provides stronger preference alignment than both supervised adaptation and standard DPO variants.

\input{table/online_vs_offline}

\section{More Qualitative Results}\label{app:more-pic-result}

\begin{figure}[h]
  \centering
  \includegraphics[width=\textwidth]{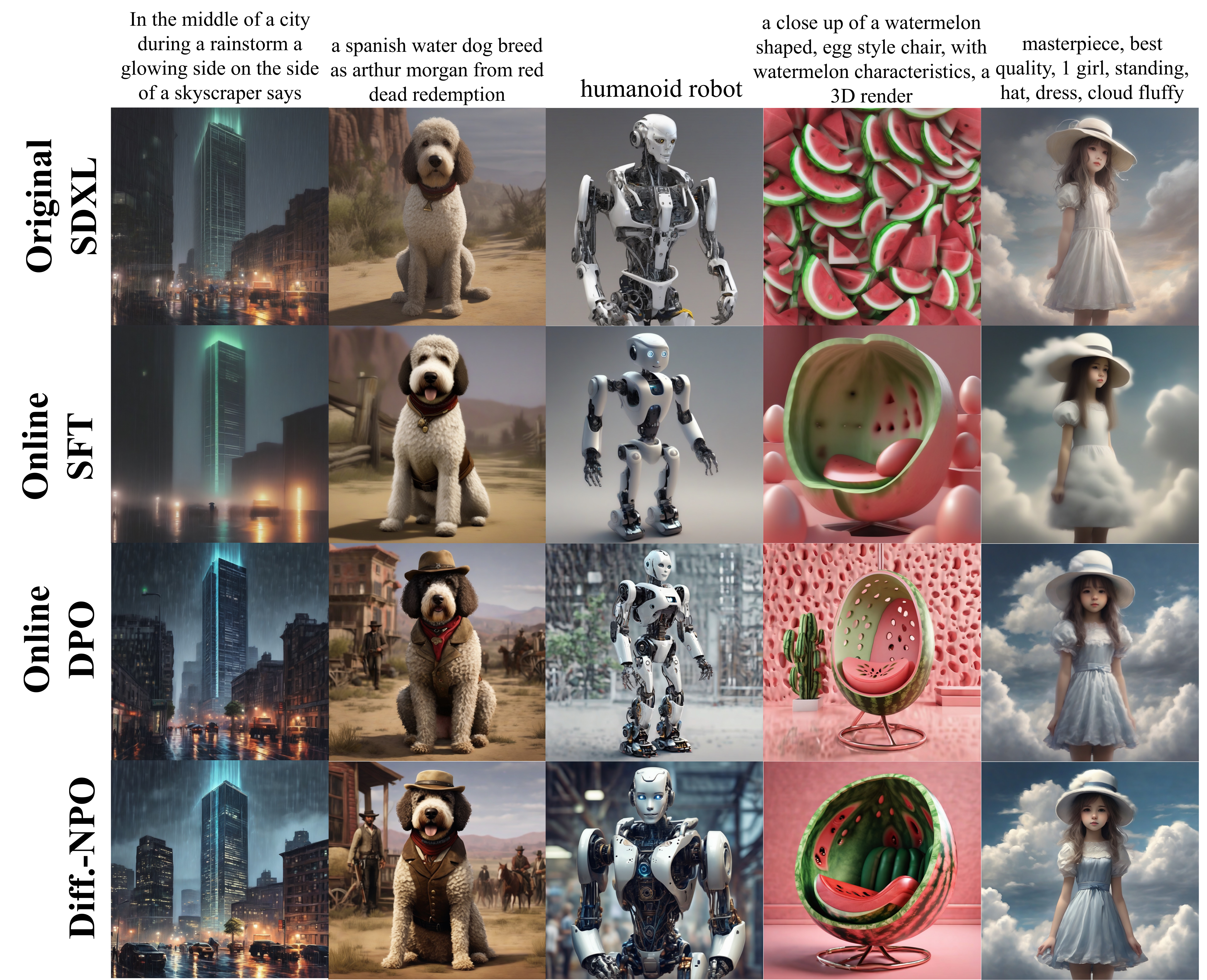}
  \caption{Additional qualitative comparison on SDXL. Compared with the baselines, \ourmech produces images with stronger prompt alignment, more coherent global structure, and improved visual quality across diverse prompts.}
  \label{fig:sdxl-pics-more}
\end{figure}

\begin{figure}[h]
  \centering
  \includegraphics[width=0.9\textwidth]{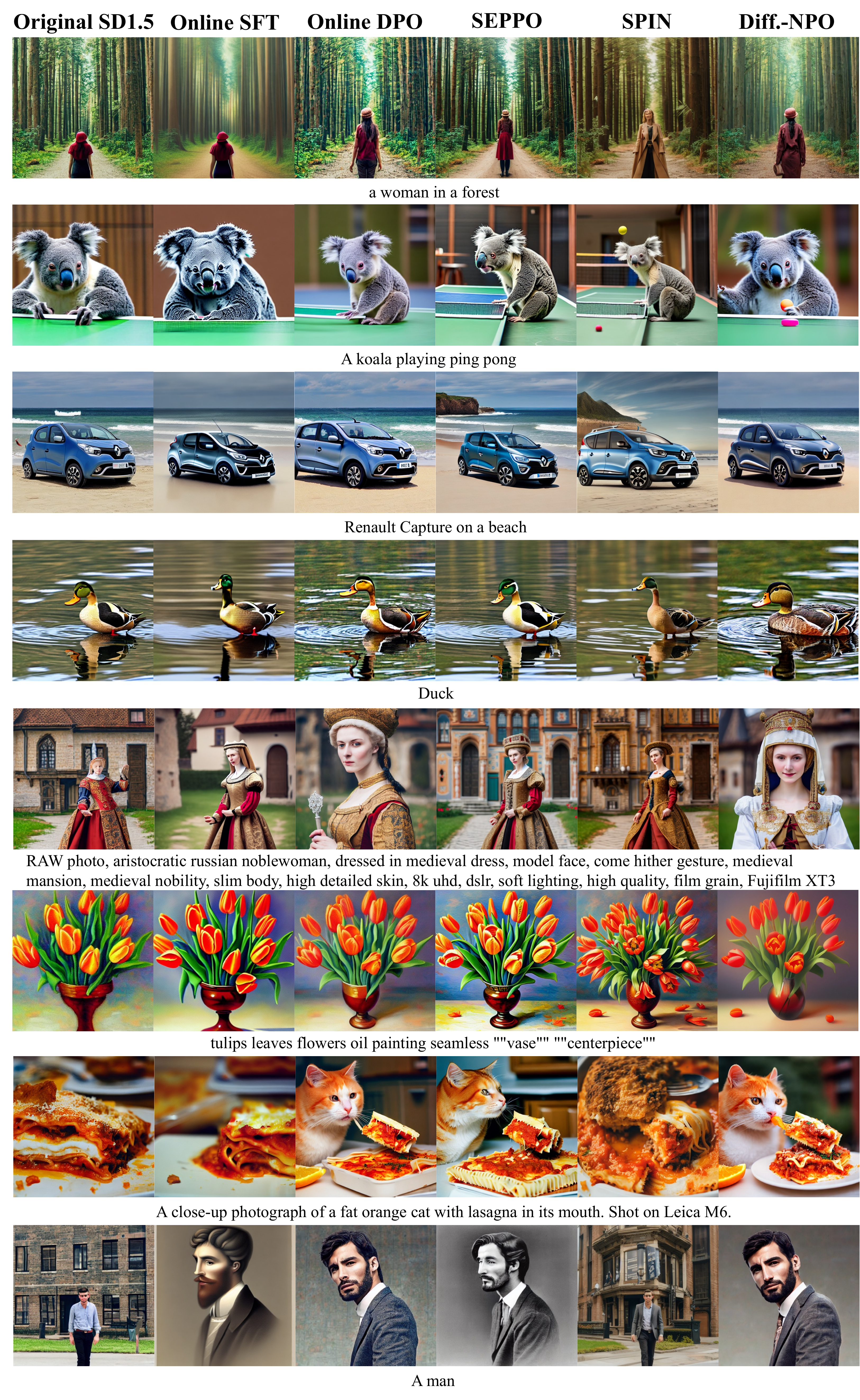}
  \caption{Qualitative comparison on SD1.5. \ourmech consistently improves the realism, semantic faithfulness, and visual coherence of generated images, demonstrating that the benefits of our preference optimization method also transfer to the SD1.5 backbone.}
  \label{fig:sd15-pics}
\end{figure}

In this section, we provide additional qualitative comparisons for both SDXL and SD1.5. As shown in Figures~\ref{fig:sdxl-pics-more} and~\ref{fig:sd15-pics}, \ourmech consistently generates images with better semantic alignment, stronger visual coherence, and more appealing overall quality compared with the baselines. These results further support the quantitative improvements reported in the main text.

%% file: table/algo.tex
\begin{algorithm}[H]
\caption{Diffusion-NPO}\label{algo:npo}
\begin{algorithmic}[1]
\State \textbf{Input:} steps $S$, KL regularization $\tau$, OMD parameter $\eta$, reference policy $p_{\mathrm{ref}}$, prompts $\rvc\sim d_0$
\State Initialize $p_1 \leftarrow p_{\mathrm{ref}}$
\For{$s=1,2,\dots,S$}
  \State Sample a minibatch of prompts $\{\rvc^{(i)}\}_{i=1}^n\sim d_0$
  \State For each $\rvc^{(i)}$, draw a batch of size $k$, denoted by $(\rvx_1^{(i)}, \ldots, \rvx_k^{(i)})\sim p_s(\cdot\mid \rvc^{(i)})$
  \State Query the average ranking oracle to obtain scores $(r_1^{(i)}, \ldots, r_k^{(i)})$ for $(\rvx_1^{(i)}, \ldots, \rvx_k^{(i)})$
  \State Rank the samples by their oracle scores and set
  \Statex \hspace{1.8em}$\rvx^{+, (i)} \leftarrow \arg\max_{\rvx_j^{(i)}} r_j^{(i)}$ and $\rvx^{-,(i)} \leftarrow \arg\min_{\rvx_j^{(i)}} r_j^{(i)}$
  \State Form the preference pairs $\{(\rvc^{(i)},\rvx^{+,(i)},\rvx^{-,(i)})\}_{i=1}^n$
  \State Update the diffusion policy by minimizing the loss with objective as equation~\ref{eq:obj-final}:
  \Statex \hspace{1.8em}$\displaystyle p_{s+1}\leftarrow \arg\min_{p\in\mathcal{P}}\; L_{\ourmech}$
  \State Implement soft update for the target policy:
  \Statex \hspace{1.8em}$\displaystyle p_{s+1}\leftarrow \lambda p_{s} + (1-\lambda)p_{s+1}$ where $\lambda$ = $\min(0.001*s, 0.5)$
\EndFor
\State \textbf{Output:} $p_{S+1}$
\end{algorithmic}
\end{algorithm}

%% file: 2_related_work.tex
\subsection{Nash Learning with Human Feedback} 
In the Nash learning with human feedback paradigm, a well-aligned policy is viewed as an optimal strategy that cannot be exploited by competing policies, thereby achieving strategic optimality rather than mere mediocrity. NLHF\citep{munos2024nash} first introduced this game-theoretic perspective by formulating preference alignment as regularized Nash equilibrium learning under general, potentially non-transitive preferences. Subsequent work has explored this paradigm through no-regret learning \citep{zhang2024iterative}, optimistic mirror descent \citep{zhang2025improving}, and extragradient updates\citep{zhou2025extragradient}, thereby advancing both theoretical guarantees and empirical stability relative to traditional RLHF. More recently, Multiplayer Nash Preference Optimization (MNPO)\citep{wu2026multiplayernashpreferenceoptimization} extends this line of work from two-player interactions to the multiplayer setting, enabling the modeling of richer preference structures.

\subsection{Diffusion Model Alignment} 

Diffusion models~\citep{SD15, SDXL} have achieved remarkable success in T2I generation. Recently, alignment methods inspired by RLHF have been increasingly adopted for diffusion models, both in direct fine-tuning~\citep{dai2023emu,  betker2023improving, clark2023directly, prabhudesai2023aligning, fan2023dpok, black2023training} and at inference time~\citep{tang2024tuning, yeh2024training, kimtest, zhai2025mira, jin2025inference}, with the goal of improving downstream reward-related objectives~\citep{flowgrpo, diffusiondpo, xue2025dancegrpo}. Existing diffusion alignment methods can be broadly grouped into three lines: (1) methods that optimize explicit reward functions, e.g., GRPO-based approaches~\citep{flowgrpo, xue2025dancegrpo, li2025uniworld, li2026aegpo}; (2) methods that optimize implicit rewards induced from pairwise preferences, such as Diffusion-DPO~\citep{diffusiondpo, liang2024step,wang2025diffusion, lee2025calibrated, gu2024diffusion, D3PO}; and (3) likelihood-free methods that optimize forward scores and velocity fields directly~\citep{dspo, diffusionNFT}. Despite their differences, most of these methods primarily focus on reward maximization, whereas only a few exploit on-policy self-play. In particular, ~\citet{D3PO} extends DPO to diffusion models by formulating denoising as a multi-step MDP and directly optimizing human preferences over sampled image pairs. ~\citet{SPIN} introduces a self-play fine-tuning scheme in which the current model learns to outperform samples generated by earlier checkpoints. ~\citet{SEPPO} reduces reliance on human-labeled losing images by replacing them with reference samples generated from previous checkpoints. ~\citet{SAIL} further constructs synthetic preference pairs by self-ranking on-policy generated images and then applies DPO-style optimization. In our work, we build on nash learning~\citep{zhang2024iterative, wu2026multiplayernashpreferenceoptimization} and develop a more comprehensive game-theoretic self-play framework for diffusion model alignment.



%% file: table/online_vs_offline.tex
\begin{table}[h]
  \centering
\small
\renewcommand{\arraystretch}{0.97}
\setlength{\tabcolsep}{2pt}
\caption{Comparison between online and offline training variants on SD1.5 Pick-a-Pic, evaluated by win rate (\%) against the original model. }
\label{tab:offline}
\begin{tabular}{lccccc}
\toprule
\multicolumn{6}{c}{\textbf{Win Rate (\%) on Pick-a-Pic}} \\
\midrule
 & PS $\uparrow$ & HPS $\uparrow$ & CLIP $\uparrow$ & IM $\uparrow$ & AES $\uparrow$ \\
\midrule
SFT(online) & 40.4 & 46.6 & 55.0 & 53.2 & 59.3 \\
SFT(offline) & 73.4 & 80.7 & 56.3 & 74.9 & 71.7 \\
DPO(online) & 77.4 & 82.5 & 60.0 & 81.0 & 73.4 \\
DPO(offline) & 73.3 & 69.8 & 57.1 & 61.7 & 63.4 \\
Diff.-NPO & \textbf{81.6} & \textbf{84.6} & \textbf{60.3} & \textbf{85.1} & \textbf{80.8} \\
\bottomrule
\end{tabular}  
\end{table}